\documentclass{article}

\usepackage{fullpage}
\usepackage{authblk}

\usepackage[utf8]{inputenc} 
\usepackage[T1]{fontenc}    
\usepackage{lmodern}
\usepackage{hyperref}       
\usepackage{url}            
\usepackage{booktabs}       
\usepackage{amsfonts}       
\usepackage{nicefrac}       
\usepackage{microtype}      
\usepackage{multicol}
\usepackage{natbib}
\usepackage{tikz}
\usepackage{diagbox}
\usepackage{multirow}
\usepackage{subfigure}
\usepackage{bbm}
\usepackage{amsmath}
\usepackage{amsthm}
\usepackage{amssymb}
\usepackage{bm}
\usepackage{footnote}
\usepackage{wrapfig}
\usepackage[export]{adjustbox}
\usepackage{notation}
\usepackage{researchpack}
\usepackage{natbib}
\usepackage[capitalise,nameinlink]{cleveref}
\usepackage{algorithm}
\usepackage[noend]{algorithmic}
\usepackage{tikz}
\usetikzlibrary{tikzmark}
\allowdisplaybreaks[4]
\usepackage[export]{adjustbox}

\crefname{defn}{Def.}{Def.}
\crefname{section}{Sec.}{Sec.}
\crefname{algorithm}{Alg.}{Alg.} 
\crefname{thm}{Thm.}{Thm.}
\crefname{lem}{Lem.}{Lem.}
\crefname{prop}{Prop.}{Prop.}

\newcommand\NoDo{\renewcommand\algorithmicdo{}}
\newcommand\ReDo{\renewcommand\algorithmicdo{\textbf{do}}}
\newcommand\ForEach{\renewcommand\algorithmicfor{\textbf{foreach}}}
\newcommand\ForOnly{\renewcommand\algorithmicfor{\textbf{for}}}

\renewcommand{\Pr}{P}

\title{
Tractable Regularization of Probabilistic Circuits
}

\author{Anji Liu$^{1}$}
\author{Guy Van den Broeck$^{1}$\vspace{5pt}}

\affil{%
  {$^{1}$Computer Science Department, University of California, Los Angeles, USA \hfill\texttt{\{liuanji|guyvdb\}@cs.ucla.edu}}
}

\begin{document}

\maketitle
 
\begin{abstract}
Probabilistic Circuits (PCs) are a promising avenue for probabilistic modeling. They combine advantages of probabilistic graphical models (PGMs) with those of neural networks (NNs). 
Crucially, however, they are tractable probabilistic models, supporting efficient and exact computation of many probabilistic inference queries, such as marginals and MAP. 
Further, since PCs are structured computation graphs, they can take advantage of deep-learning-style parameter updates, which greatly improves their scalability.
However, this innovation also makes PCs prone to overfitting, which has been observed in many standard benchmarks.
Despite the existence of abundant regularization techniques for both PGMs and NNs, they are not effective enough when applied to PCs.
Instead, we re-think regularization for PCs and propose two intuitive techniques, \emph{data softening} and \emph{entropy regularization}, that both take advantage of PCs' tractability and still have an efficient implementation as a computation graph. 
Specifically, data softening provides a principled way to add uncertainty in datasets in closed form, which implicitly regularizes PC parameters. 
To learn parameters from a softened dataset, PCs only need linear time by virtue of their tractability.
In entropy regularization, the exact entropy of the distribution encoded by a PC can be regularized directly, which is again infeasible for most other density estimation models.
We show that both methods consistently improve the generalization performance of a wide variety of PCs. Moreover, when paired with a simple PC structure, we achieved state-of-the-art results on 10 out of 20 standard discrete density estimation benchmarks.

\end{abstract}

\section{Introduction}
\label{sec:intro}

Probabilistic Circuits (PCs) \citep{choiprobabilistic,dang2021juice} are considered to be the lingua franca for Tractable Probabilistic Models (TPMs) as they offer a unified framework to abstract from a wide variety of TPM circuit representations, such as arithmetic circuits (ACs) \citep{darwiche2003differential}, sum-product networks (SPNs) \citep{poon2011sum}, and probabilistic sentential decision diagrams (PSDDs) \citep{kisa2014probabilistic}. PCs are a successful combination of classic probabilistic graphical models (PGMs) and neural networks (NNs). 
Moreover, by enforcing various structural properties, PCs permit efficient and exact computation of a large family of probabilistic inference queries \citep{vergari2021compositional,KhosraviNeurIPS19,shen2016tractable}. The ability to answer these queries leads to successful applications in areas such as model compression \citep{LiangXAI17} and model bias detection \citep{choi2020group,choi2020learning}. 
At the same time, PCs are analogous to NNs since their evaluation is also carried out using computation graphs. By exploiting the parallel computation power of GPUs, dedicated implementations \citep{dang2021juice,molina2019spflow} can train a complex PC with millions of parameters in minutes. These innovations have made PCs much more expressive and scalable to richer datasets that are beyond the reach of ``older'' TPMs \citep{peharz2020einsum}.

However, such advances make PCs more prone to overfitting. Although parameter regularization has been extensively studied in both the PGM and NN communities \citep{srivastava2014dropout,ioffe2015batch}, we find that existing regularization techniques for PGMs and NNs are either not suitable or not effective enough when applied to PCs. For example, parameter priors or Laplace smoothing typically used in PGMs, and often used in PC learning as well \citep{liang2017learning,dang2020strudel,gens2013learning}, incur unwanted bias when learning PC parameters -- we will illustrate this point in \cref{sec:pc}. Classic NN methods such as L1 and L2 regularization are not always suitable since PCs often use either closed-form or EM-based parameter~updates.

This paper designs parameter regularization methods that are directly tailored for PCs. We propose two regularization techniques, \emph{data softening} and \emph{entropy regularization}. Both formulate the regularization objective in terms of distributions, regardless of their representation and parameterization. Yet, both leverage the tractability and structural properties of PCs. Specifically, data softening injects noise into the dataset by turning hard evidence in the samples into soft evidence \citep{chan2005revision,pan2006belief}. While learning with such softened datasets is infeasible even for simple machine learning models, with their tractability, PCs can learn the maximum-likelihood estimation (MLE) parameters given a softened dataset in $\bigO(\abs{\p} \!\cdot\! \abs{\data})$ time, where $\abs{\p}$ is the size of the PC and $\abs{\data}$ is the size of the (original) dataset.
Additionally, the entropy of the distribution encoded by a PC can be tractably regularized. Although the entropy regularization objective for PC is multi-modal and a global optimum cannot be found in general, we propose an algorithm that is guaranteed to converge monotonically towards a stationary point.

We show that both proposed approaches consistently improve the test set performance over standard density estimation benchmarks. Furthermore, we observe that when data softening and entropy regularization are properly combined, even better generalization performance can be achieved. Specifically, when paired with a simple PC structure, this combined regularization method achieves state-of-the-art results on 10 out of 20 standard discrete density estimation benchmarks.

\boldparagraph{Notation} We denote random variables by uppercase letters (e.g., $X$) and their assignments by lowercase letters (e.g., $x$). Analogously, we use bold uppercase letters (e.g., $\X$) and bold lowercase letters (e.g., $\x$) for sets of variables and their joint assignments, respectively.

\section{Two Intuitive Ideas for Regularizing Distributions}
\label{sec:intuitive-ideas}

A common way to prevent overfitting in machine learning models is to regularize the syntactic representation of the distribution. For example, L1 and L2 losses add mutually independent priors to all parameters of a model; other approaches such as Dropout \citep{srivastava2014dropout} and Bayesian Neural Networks (BNNs) \citep{goan2020bayesian} incorporate more complex and structured priors into the model \citep{gal2016dropout}. 
In this section, we ask the question: how would we regularize an arbitrary distribution, regardless of the model at hand, and the way it is parameterized?
Such global, model-agnostic regularizers appear to be under-explored. Next, we introduce two intuitive ideas for regularizing distributions, and study how they can be practically realized in the context of probabilistic circuits in the remainder of this paper.


\boldparagraph{Data softening} 
Data augmentation is a common  technique to improve the generalization performance of machine learning models \citep{perez2017effectiveness,szegedy2016rethinking}. A simple yet effective type of data augmentation is to inject noise into the samples, for example by randomly corrupting bits or pixels \citep{vincent2008extracting}. This can greatly improve generalization as it renders the model more robust to such noise. 
While current noise injection methods are implemented as a sequence of sampled transformations, we stress that some noise injection can be done in closed form: we will be considering all possible corruptions, each with their own probability, as a function of how similar they are to a training data point.

Consider boolean variables\footnote{We postpone the discussion on regularizing samples with non-boolean variables in \cref{sec:softreg-non-boolean}.} as an example: after noise injection, a sample $X \!=\! 1$ is represented as a distribution over all possible assignments (\ie $X\!=\!1$ and $X\!=\!0$), where the instance $X\!=\!1$, which is ``similar'' to the original sample, gets a higher probability: $\Pr(X\!=\!1) \!=\! \beta$. Here $\beta \!\in\! (0.5, 1]$ is a hyperparameter that specifies the regularization strength --- if $\beta \!=\! 1$, no regularization is added; if $\beta$ approaches $0.5$, the regularized sample represents an (almost) uniform distribution. 
For a sample $\x$ with $K$ variables $\X\!:=\!\{X_i\}_{i=1}^{K}$, where the $k$th variable takes value $x_{k}$, we can similarly `soften' $\x$ by independently injecting noise into each variable, resulting in a \emph{softened distribution} $\Pr_{\x,\beta}$:
    \begin{align*}
        \forall \x' \!\in\! \val(\X), \quad \Pr_{\x,\beta} (\X \!=\! \x') := \prod_{i=1}^{K} \Pr_{\x,\beta}(X_i \!=\! x'_i) = \prod_{i=1}^{K} \Big ( \beta \!\cdot\! \indicator{x'_i \!=\! x_i} + (1\!-\!\beta) \!\cdot\! \indicator{x'_i \!\neq\! x_i} \Big ).
    \end{align*}
\noindent 
For a full dataset $\data \!:=\! \{\x^{(i)}\}_{i=1}^{N}$, this softening of the data can also be represented through a new, \emph{softened dataset} $\data_{\beta}$. Its empirical distribution is the average softened distribution of its data. It is a weighted dataset, where $\weight{\data_{\beta}}{\x}$ denotes the weight of sample $\x$ in $\data_{\beta}$:
    \begin{align}
        \data_{\beta} := \{\x \!\mid\! \x \!\in\! \val(\X)\} \quad \text{ and } \quad \weight{\data_{\beta}}{\x} = \frac{1}{N} \sum_{i=1}^{N} \Pr_{\x^{(i)},\beta} (\X = \x). \label{eq:soft-dataset}
    \end{align}
This softened dataset ensures that each possible assignment has a small but non-zero weight in the training data. Consequently, any distribution learned on the softened data must assign a small probability everywhere as well.
Of course, materializing this dataset, which contains all possible training example, is not practical. Regardless, we will think of data softening as implicitly operating on this softened dataset.
We remark that data softening is related to soft evidence \citep{jeffrey1990logic} and virtual evidence \citep{pearl2014probabilistic}, which both define a framework to incorporate uncertain evidence into a distribution.

\boldparagraph{Entropy regularization}
Shannon entropy is an effective indicator for overfitting. For a dataset $\data$ with $N$ distinct samples, a perfectly overfitting model that learns the exact empirical distribution has entropy $\log(N)$. A distribution that generalizes well should have a much larger entropy, since it assigns positive probability to exponentially more assignments near the training samples. Concretely, for the protein sequence density estimation task \citep{russ2020evolution} that we will experiment with in \cref{sec:new-reg-methods-nondet}, the perfectly overfitting empirical distribution has entropy $3$, a severely overfitting learned model has entropy $92$, yet a model that generalizes well has entropy $177$. Therefore, directly controlling the entropy of the learned distribution will help mitigate overfitting. 
Given a model $\Pr_{\params}$ parametrized by $\params$ and a dataset $\data \!:=\! \{\x^{(i)}\}_{i=1}^{N}$, we define the following entropy regularization objective:
    \begin{align}
        \LL_{\mathrm{ent}}(\params; \data, \tau) := \frac{1}{N} \sum_{i=1}^{N} \log \Pr_{\params} (\x^{(i)}) + \tau \cdot \entropy(\Pr_{\params}),
        \label{eq:det-pc-ent-reg-mle}
    \end{align}
\noindent where $\entropy(\Pr_{\params}) \!:=\! -\sum_{\x\in\val(\X)} \Pr_{\params} (\x) \log \Pr_{\params} (\x)$ denotes the entropy of distribution $\Pr_{\params}$, and $\tau$ is a hyperparameter that controls the regularization strength. Various forms of entropy regularization have been used in the training process of deep learning models. Different from \cref{eq:det-pc-ent-reg-mle}, these methods regularize the entropy of a parametric \citep{grandvalet2006entropy,zhu2017unpaired} or non-parametric \citep{feng2017learning} output space of the model.

Although both ideas for regularizing distributions are rather intuitive, it is surprisingly hard to implement them in practice since they are intractable even for the simplest machine learning models.
\begin{thm}
\label{thm:hardness-soft-reg}
Computing the likelihood of a distribution represented as a exponentiated logistic regression (or equivalently, a single neuron) given softened data is \#P-hard.
\end{thm}

\begin{thm}
\label{thm:hardness-ent-reg}
Computing the Shannon entropy of a normalized logistic regression model is \#P-hard.
\end{thm}

Proof of \cref{thm:hardness-soft-reg,thm:hardness-ent-reg} are provided in \cref{sec:proof-hardness-soft-reg,sec:proof-sharpp-hardness-ent-reg}. Although data softening and entropy regularization are infeasible for many models, we will show in the following sections that they are tractable to use when applied to Probabilistic Circuits (PCs) \citep{choiprobabilistic}, a class of expressive TPMs.



\section{Background and Motivation}
\label{sec:pc}

Probabilistic Circuits (PCs) are a collective term for a wide variety of TPMs. They present a unified set of notations that provides succinct representations for TPMs such as Probabilistic Sentential Decision Diagrams (PSDDs) \citep{kisa2014probabilistic}, Sum-Product Networks (SPNs) \citep{poon2011sum}, and Arithmetic Circuits (ACs) \citep{darwiche2003differential}. We proceed by introducing the syntax and semantics of a PC.


\begin{defn}[Probabilistic Circuits]
\label{def:pc}
A PC $\p$ that represents a probability distribution over variables $\X$ is defined by a parametrized directed acyclic graph (DAG) with a single root node, denoted $n_r$. The DAG comprises three kinds of units: \emph{input}, \emph{sum}, and \emph{product}. Each leaf node $n$ in the DAG corresponds to an input unit; each inner node $n$ (\ie sum and product units) receives inputs from its children, denoted $\ch(n)$. Each unit $n$ encodes a probability distribution $\p_n$, defined as follows:
    \begin{align*}
        \p_n (\x) := \begin{cases}
            f_n (\x) & \text{if}~n~\text{is~an~input~unit}, \\
            \sum_{c\in\ch(n)} \theta_{n, c} \cdot \p_c (\x) & \text{if}~n~\text{is~a~sum~unit}, \\
            \prod_{c\in\ch(n)} \p_c (\x) & \text{if}~n~\text{is~a~product~unit},
        \end{cases}
    \end{align*}
\noindent where $f_n$ is a univariate input distribution (e.g., boolean, categorical or Gaussian), and $\theta_{n, c}$ represents the parameter corresponds to edge $(n, c)$. Intuitively, a sum unit models a weighted mixture distribution over its children, and a product unit encodes a factored distribution over its children. We assume w.l.o.g.\ that all parameters are positive and the parameters associated with any sum unit $n$ sum up to 1 (\ie $\sum_{c\in\ch(n)} \theta_{n,c} \!=\! 1$). We further assume w.l.o.g.\ that a PC alternates between sum and product layers \citep{vergari2015simplifying}. The size of a PC $\p$, denoted $\abs{\p}$, is the number of edges in its DAG.
\end{defn}

This paper focuses on two classes of PCs that support different types of queries: (i) PCs that allow linear-time computation of marginal (MAR) and maximum-a-posterior (MAP) inferences (\eg PSDDs \citep{kisa2014probabilistic}, selective SPNs \citep{peharz2014learning}); (ii) PCs that only permit linear-time computation of MAR queries (\eg SPNs \citep{poon2011sum}). The borders between these two types of PCs are defined by their \emph{structural properties}, \ie constraints imposed on a PC. First, in order to compute MAR queries in linear time, both classes of PCs should be decomposable (\cref{def:decomposability}) and smooth (\cref{def:smoothness}) \citep{choiprobabilistic}.
These are properties of the (variable) scope $\phi(n)$ of PC units $n$, that is, the collection of variables defined by all its descendent input nodes.

\begin{defn}[Decomposability]
\label{def:decomposability}
A PC is decomposable if for every product unit $n$, its children have disjoint scopes: $\forall c_1, c_2 \in \ch(n) \; (c_1 \neq c_2), \phi(c_1) \cap \phi(c_2) = \emptyset$.
\end{defn}

\begin{defn}[Smoothness]
\label{def:smoothness}
A PC is smooth if for every sum unit $n$, its children have the same scope: $\forall c_1, c_2 \in \ch(n), \phi(c_1) = \phi(c_2)$. 
\end{defn}

Next, \emph{determinism} is required to guarantee efficient computation of MAP inference \citep{mei2018maximum}.


\begin{defn}[Determinism]
\label{def:determinism}
Define the support $\supp(n)$ of a PC unit $n$ as the set of complete variable assignments $\x\in\val(\X)$ for which $\p_n (\x)$ has non-zero probability (density): $\supp(n) = \{ \x \mid \x\!\in\!\val(\X), \p_n (\x) \!>\! 0\}$. A PC is deterministic if for every sum unit $n$, its children have disjoint support: $\forall c_1, c_2 \in \ch(n) \; (c_1 \neq c_2), \supp(c_1) \cap \supp(c_2) = \emptyset$.
\end{defn}

Since the only difference in the structural properties of both PCs classes is determinism, we denote members in the first PC class as deterministic PCs, and members in the second PC class as non-deterministic PCs. Interestingly, both PC classes not only differ in their tractability, which is characterized by the set of queries that can be computed within $\poly{\abs{\p}}$ time \citep{vergari2021compositional}, they also exhibit drastically different expressive efficiency. Specifically, abundant empirical \citep{dang2020strudel,peharz2020einsum} and theoretical \citep{choi2017relaxing} evidences suggest that non-deterministic PCs are more expressive than their deterministic counterparts. Due to their differences in terms of tractability and expressive efficiency, this paper studies parameter regularization on deterministic and non-deterministic PCs separately.


\begin{figure}[t]
    \centering
    \includegraphics[width=0.85\columnwidth]{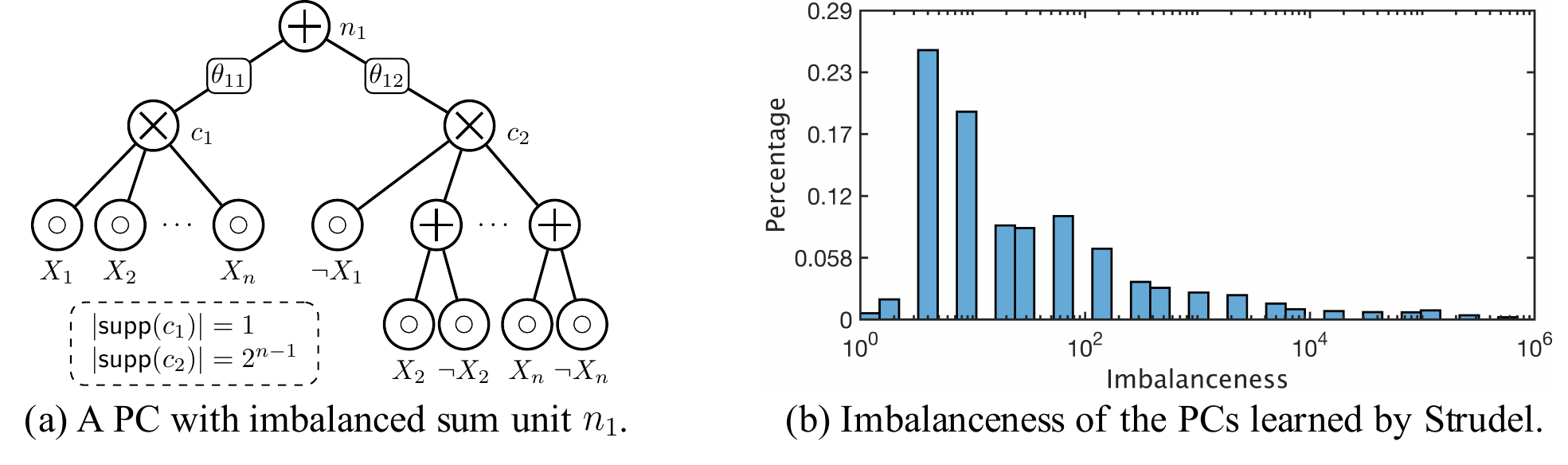}
    \vspace{-1.2em}
    \caption{A Problem of Laplace smoothing. (a) Laplace smoothing cannot properly regularize this PC as the sum unit $n_1$ is imbalanced, \ie its two children have drastically different support sizes. (b) A large fraction of sum units learned by a PC structure learning algorithm \citep{dang2020strudel} are imbalanced.}
    \label{fig:pseudocount-problems}
\end{figure}

\boldparagraph{Motivation} 
Laplace smoothing is widely adopted as a PC regularizer \citep{liang2017learning,dang2020strudel}. Since it is also the default regularizer for classical probabilistic models such as Bayesian Networks (BNs) \citep{heckerman2008tutorial} and Hierarchical Bayesian Models (HBMs) \citep{allenby2006hierarchical}, this naturally raises the following question: \emph{are there differences between a good regularizer for classical probabilistic models such as BNs and HBMs and effective regularizers for PCs?} The question can be answered affirmatively --- while Laplace smoothing provides good priors to BNs and HBMs, its uniform prior could add unwanted bias to PCs. Specifically, for every sum unit $n$, Laplace smoothing assigns the same prior to all its child parameters (\ie $\{\theta_{n,c} \mid c\!\in\!\ch(n)\}$), while in many practical PCs, these parameters should be given drastically different priors. For example, consider the PC shown in \cref{fig:pseudocount-problems}(a). Since $c_2$ has an exponentially larger support than $c_1$, it should be assumed as prior that $\theta_{12}$ will be much larger than $\theta_{11}$.



We highlight the significance of the above issue by examining the fraction of sum units with imbalanced child support sizes in PCs learned by Strudel, a state-of-the-art structure learning algorithm for deterministic PCs \citep{kisa2014probabilistic}. We examine 20 PCs learned from the 20 density estimation benchmarks \citep{van2012markov}, respectively. All sum units with $\geq\!3$ children and with a support size $\geq\!128$ are recorded. We measure ``imbalanceness'' of a sum unit $n$ by the fraction of the maximum and minimum support size of its children (i.e., $\frac{\max_{c_1\in\ch(n)} \abs{\supp(c_1)}} {\min_{c_2\in\ch(n)} \abs{\supp(c_2)}}$). As demonstrated in \cref{fig:pseudocount-problems}(b), more than $20\%$ of the sum units have imbalanceness $\geq\!10^2$, which suggests that the inability of Laplace smoothing to properly regularize PCs with imbalanced sum units could lead to severe performance degradation in practice.

\section{How Is This Tractable And Practical?}
\label{sec:tractable-reg}

In this section, we first provide additional background about the parameter learning algorithms for deterministic and non-deterministic PCs (\cref{sec:pc-learning}). We then demonstrate how the two intuitive ideas for regularizing distributions (\cref{sec:intuitive-ideas}), \ie data softening and entropy regularization, can be efficiently implemented for deterministic (\cref{sec:new-reg-methods-det}) and non-deterministic (\cref{sec:new-reg-methods-nondet}) PCs.


\begin{figure}[t]
\begin{minipage}[t]{0.50\textwidth}
\begin{algorithm}[H]
\caption{Forward pass}
\label{alg:forward-pass}
{\fontsize{9}{9} \selectfont
\begin{algorithmic}[1]

\STATE {\bfseries Input:} A deterministic PC $\p$; sample $\x$

\STATE {\bfseries Output:} $\mathtt{value}[n] \!\!:=\!\! (\x\!\in\!\supp(n))$ for each unit $n$

\ForEach
\FOR{\tikzmarknode{a1}{} $n$ traversed in postorder}
\STATE \textbf{if} $n$ \textbf{isa} input unit \textbf{then} $\mathtt{value}[n]\!\leftarrow\!f_n(\x)$
\STATE \textbf{elif} \tikzmarknode{a2}{} $n$ \textbf{isa} product unit \textbf{then}
\STATE \hspace{1em} $\mathtt{value}[n]\!\leftarrow\!\prod_{c\in\ch(n)} \mathtt{value}[c]$
\STATE \textbf{else} \tikzmarknode{a3}{} \slash\slash $n$ is a sum unit
\STATE \hspace{1em} $\mathtt{value}[n]\!\leftarrow\!\sum_{c\in\ch(n)} \mathtt{value}[c]$
\ENDFOR
\ForOnly
\end{algorithmic}
}
\end{algorithm}
\end{minipage}
\hfill
\begin{minipage}[t]{0.49\textwidth}
\begin{algorithm}[H]
\caption{Backward pass}
\label{alg:backward-pass}
{\fontsize{9}{9} \selectfont
\begin{algorithmic}[1]

\STATE {\bfseries Input:} A deterministic PC $\p$; $\forall n, \mathtt{value}[n]$

\STATE {\bfseries Output:} $\mathtt{flow}[n,c] \!:=\! (\x\!\in\!(\context_n \!\cap\! \context_c))$ for each pair $(n,c)$, where $n$ is a sum unit and $c\!\in\!\ch(n)$

\STATE $\forall n, \mathtt{context}[n]\!\leftarrow\!0$; $\mathtt{context}[n_r]\!\leftarrow\! \mathtt{value}[n_r]$

\ForEach
\FOR{\tikzmarknode{b1}{} sum unit $n$ traversed in preorder}
\NoDo
\FOR{\tikzmarknode{b2}{} $m \in \pa(n)$ \textbf{do} $\;$ (denote $g\!\leftarrow\!\pa(m)$)}
\STATE $\mathtt{f} \leftarrow \frac{\mathtt{value}[m]}{\mathtt{value}[g]} \cdot \mathtt{context}[g]$
\STATE $\mathtt{context}[n] \pluseq \mathtt{f}; \quad \mathtt{flow}[g,m] = \mathtt{f}$
\ENDFOR
\ReDo
\ENDFOR
\ForOnly

\end{algorithmic}
}
\end{algorithm}
\end{minipage}
\begin{tikzpicture}[overlay,remember picture]
   \draw[black,line width=0.6pt] ([xshift=-32pt,yshift=-3pt]a1.west) -- ([xshift=-32pt,yshift=-52pt]a1.west) -- ([xshift=-28pt,yshift=-52pt]a1.west);
   \draw[black,line width=0.6pt] ([xshift=-13pt,yshift=-3pt]a2.west) -- ([xshift=-13pt,yshift=-12pt]a2.west) -- ([xshift=-9pt,yshift=-12pt]a2.west);
   \draw[black,line width=0.6pt] ([xshift=-14.4pt,yshift=-3pt]a3.west) -- ([xshift=-14.4pt,yshift=-12pt]a3.west) -- ([xshift=-10.4pt,yshift=-12pt]a3.west);
   \draw[black,line width=0.6pt] ([xshift=-32pt,yshift=-3pt]b1.west) -- ([xshift=-32pt,yshift=-37pt]b1.west) -- ([xshift=-28pt,yshift=-37pt]b1.west);
   \draw[black,line width=0.6pt] ([xshift=-32pt,yshift=-3pt]b2.west) -- ([xshift=-32pt,yshift=-27pt]b2.west) -- ([xshift=-28pt,yshift=-27pt]b2.west);
\end{tikzpicture}
\end{figure}

\subsection{Learning the Parameters of PCs}
\label{sec:pc-learning}

\boldparagraph{Deterministic PCs}
Given a deterministic PC $\p$ defined on variables $\X$ and a dataset $\data = \{\x^{(i)}\}_{i=1}^{N}$, the maximum likelihood estimation (MLE) parameters $\params^{*}_{\data} \!:=\! \argmax_{\params} \sum\nolimits_{i=1}^{N} \log \p(\x^{(i)} ; \params)$ can be learned in closed-form. To formalize the MLE solution, we need a few extra definitions.

\begin{defn}[Context]
\label{def:context}
The context $\context_n$ of every unit $n$ in a PC $\p$ is defined in a top-down manner: for the base case, context of the root node $n_r$ is defined as its support: $\context_{n_r} := \supp(n_r)$. For every other node $n$, its context is the intersection of its support and the union of its parents' ($\pa(n)$) contexts:
    \begin{align*}
        \context_n := \bigcup_{m\in\pa(n)} \context_m \cap \supp(n).
    \end{align*}
\end{defn}

Intuitively, if an assignment $\x$ is in the context of unit $n$, then there exists a path on the PC's DAG from $n$ to the root unit $n_r$ such that for any unit $m$ in the path, we have $\x \!\in\! \supp(m)$. Circuit flow extends the notation of context to indicate whether a sample $\x$ is in the context of an edge $(n,c)$.

\begin{defn}[Flows]
\label{def:flows}
The flow $\flow_{n,c} (\x)$ of any edge $(n,c)$ in a PC given variable assignments $\x\!\in\!\val(\X)$ is defined as $\indicator{\x\!\in\!\context_n\!\cap\!\context_c}$, where $\indicator{\cdot}$ is the indicator function. The flow $\flow_{n,c} (\data)$ w.r.t. dataset $\data \!=\! \{\x^{(i)}\}_{i=1}^{N}$ is the sum of the flows of all its samples: $\flow_{n,c} (\data) \!:=\! \sum_{i=1}^{N} \flow_{n,c} (\x^{(i)})$.
\end{defn}

The flow $\flow_{n,c} (\x)$ for \emph{all} edges $(n,c)$ in a PC $\p$ w.r.t. sample $\x$ can be computed through a forward and backward path that both take $\bigO(\abs{\p})$ time. The forward path, as shown in \cref{alg:forward-pass}, starts from the leaf units and traverses the PC in postorder to compute $\forall n, \mathtt{value}[n] \!:=\! \indicator{\x\!\in\!\supp(n)}$; afterwards, the backward path illustrated in \cref{alg:backward-pass} begins at the root unit $n_r$ and traverses the PC in preorder to compute $\forall n, \mathtt{context}[n] \!:=\! \indicator{\x\!\in\!\context_n}$ as well as $\forall (n,c), \mathtt{flow}[n,c] \!:=\! \flow_{n,c} (\x)$. By \cref{def:flows}, the time complexity for computing $\flow_{n,c} (\data)$ with respect to all edges $(n,c)$ in $\p$ is $\bigO(\abs{\p} \!\cdot\! \abs{\data})$, where $\abs{\data}$ is the size of dataset $\data$. The correctness of \cref{alg:forward-pass,alg:backward-pass} are justified in \cref{sec:correctness-f-b}.

The MLE parameters $\params^{*}_{\data}$ given dataset $\data$ can be computed using the flows \citep{kisa2014probabilistic}:
    \begin{align}
        \forall (n,c), \quad \theta^{*}_{n,c} = \flow_{n,c} (\data) / \sum\nolimits_{c\in\ch(n)} \flow_{n,c} (\data).
        \label{eq:det-pc-mle-params}
    \end{align}
Define hyperparameter $\pseudocount$ ($\pseudocount \!\geq\! 0$), for every sum unit $n$, Laplace smoothing regularizes its child parameters (\ie $\{\theta_{n,c} \!\mid\! c \!\in\! \ch(n)\}$) by adding a \emph{pseudocount} $\pseudocount / \abs{\ch(n)}$ to every child branch of $n$, which is equivalent to adding $\pseudocount / \abs{\ch(n)}$ to the numerator of \cref{eq:det-pc-mle-params} and $\pseudocount$ to its denominator.
    
\begin{figure}[t]
    \centering
    \begin{minipage}{.39\textwidth}
        \includegraphics[width=\columnwidth]{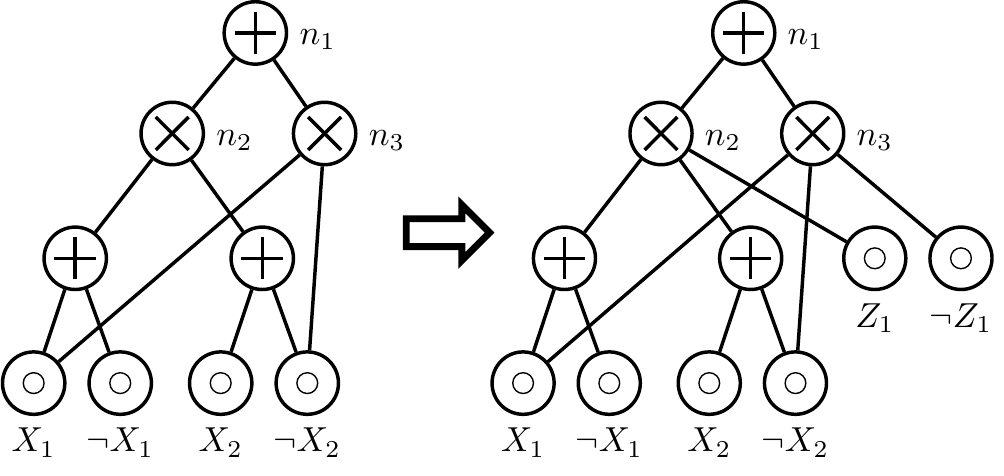}
        \vspace{-2.0em}
        \caption{A non-deterministic PC can be modified as an equivalent deterministic PC with hidden variables.
        }
        \label{fig:nondet-pc1}
    \end{minipage}
    \hspace{1em}
    \begin{minipage}{.25\textwidth}
        \includegraphics[width=\columnwidth]{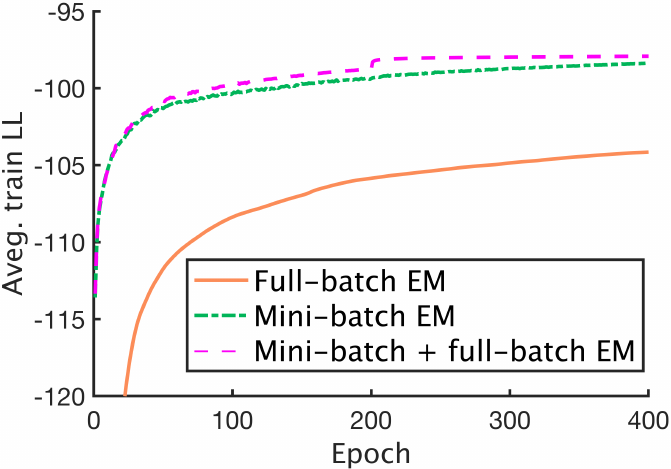}
        \vspace{-1.6em}
        \caption{Average train LL on MNIST using different EM updates.}
        \label{fig:nondet-pc2}
    \end{minipage}
    \hspace{1em}
    \begin{minipage}{.27\textwidth}
        \includegraphics[width=\columnwidth]{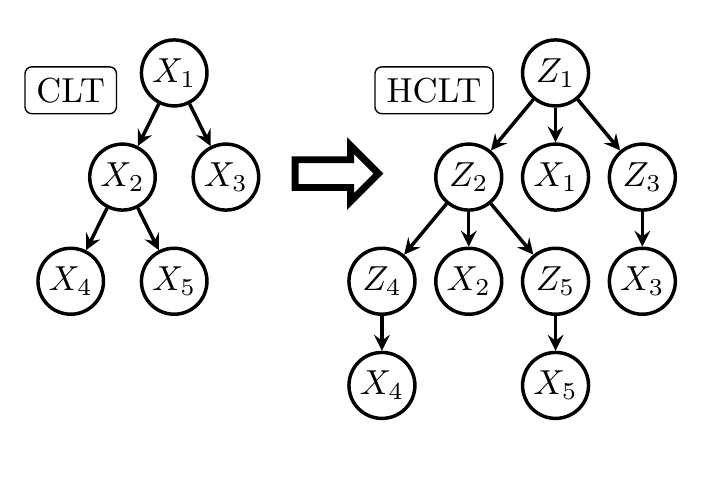}
        \vspace{-1.8em}
        \caption{HCLT is constructed by adding hidden variables in a CLT \citep{chow1968approximating}.}
        \label{fig:nondet-pc3}
    \end{minipage}
\end{figure}

    
\boldparagraph{Non-deterministic PCs}
As justified by Peharz et al. \citep{peharz2016latent}, every non-deterministic PC can be augmented as a deterministic PC with additional hidden variables. For example, in \cref{fig:nondet-pc1}, the left PC is not deterministic since the support of both children of $n_1$ (\ie $n_2$ and $n_3$) contains $x_1 \bar{x}_2$. The right PC augments the left one by adding input units correspond to hidden variable $Z_1$, which retains determinism by ``dividing'' the overlapping support $x_1 \bar{x}_2$ into $x_1 \bar{x}_2 z_1 \!\in\! \supp(n_2)$ and $x_1 \bar{x}_2 \bar{z}_1 \!\in\! \supp(n_3)$. Under this interpretation, parameter learning of non-deterministic PCs is equivalent to learning the parameters of deterministic PCs given incomplete data (we never observe the hidden variables), which can be solved by Expectation-Maximization (EM) \citep{darwiche2009modeling,dempster1977maximum}. In fact, EM is the default parameter learning algorithm for non-deterministic PCs \citep{peharz2020einsum,choi2020group}.

Under the latent variable model view of a non-deterministic PC, its EM updates can be computed using \emph{expected flows} \citep{choi2020group}. Specifically, given observed variables $\X$ and (implicit) hidden variables $\Z$, the expected flow of edge $(n,c)$ given dataset $\data$ is defined as
    \begin{align*}
        \expflow_{n,c} (\data;\params) := \expectation_{\x\sim\data, \z\sim\p_c(\cdot\mid\x;\params)} [\flow_{n,c} (\x,\z)],
    \end{align*}
\noindent where $\params$ is the set of parameters, and $\p_c(\cdot\mid\x;\params)$ is the conditional probability over hidden variables $\Z$ given $\x$ specified by the PC rooted at unit $c$. Similar to flows, the expected flows can be computed via a forward and backward pass of the PC (\cref{alg:nondet-forward-pass,alg:nondet-backward-pass} in the Appendix). As shown by Choi et al. \cite{choi2020group}, for a non-deterministic PC, its parameters for the next EM iteration are given by
    \begin{align}
        \theta^{(new)}_{n,c} = \expflow_{n,c} (\data; \params) / \sum_{c\in\ch(n)} \expflow_{n,c} (\data; \params).
        \label{eq:pc-em-update}
    \end{align}
This paper uses a hybrid EM algorithm, which uses mini-batch EM updates to initiate the training process, and switch to full-batch EM updates afterwards. Specifically, in mini-batch EM, $\params^{(new)}$ are computed using mini-batches of samples, and the parameters are updated towards the taget with a step size $\eta$: $\params^{(k+1)} \!\leftarrow\! (1-\eta) \params^{(k)} + \eta \params^{(new)}$; when using full-batch EM, we iteratively compute the updated parameters $\params^{(new)}$ using the whole dataset. \cref{fig:nondet-pc2} demonstrates that this hybrid approach offers faster convergence speed compared to using full-batch or mini-batch EM only.
    

\subsection{Regularizing Deterministic PCs}
\label{sec:new-reg-methods-det}

We demonstrate how the intuitive ideas for regularizing distributions presented in \cref{sec:intuitive-ideas} (\ie data softening and entropy regularization) can be efficiently applied to deterministic PCs.


\boldparagraph{Data softening} 
As hinted by \cref{eq:soft-dataset}, we need exponentially many samples to represent a softened dataset, which makes parameter learning intractable even for the simple logistic regression model (\cref{thm:hardness-soft-reg}), let alone more complex probabilistic models such as VAEs \citep{kingma2013auto} and GANs \citep{goodfellow2014generative}. Despite this negative result, the MLE parameters of a PC $\p$ \wrt $\data_{\beta}$ can be computed in time $\bigO(\abs{\p} \!\cdot\! \abs{\data})$, which is linear \wrt the model size as well as the size of the \emph{original} dataset.


\begin{thm}
\label{thm:data-softening-alg}
Let $f_n(\x) \!=\! \beta \!\cdot\! \indicator{\x\in\supp(n)} + (1\!-\!\beta) \!\cdot\! \indicator{\x\not\in\supp(n)}$ in \cref{alg:forward-pass}.
Given a deterministic PC $\p$, a boolean dataset $\data$, and hyperparameter $\beta \!\in\! (0.5, 1]$, the set of all flows $\{\flow_{n,c}(\data_{\beta}) \mid \forall~\textrm{edge}~(n,c)\}$ w.r.t. the softened dataset $\data_{\beta}$ can be computed by \cref{alg:forward-pass,alg:backward-pass} within $\bigO(\abs{\p} \!\cdot\! \abs{\data})$ time.
\end{thm}

Proof of this theorem is provided in \cref{sec:proof-data-soft-alg}. Since the MLE parameters (\cref{eq:det-pc-mle-params}) \wrt $\data_{\beta}$ can be computed in $\bigO(\abs{\p})$ time using the flows, the overall time complexity to compute the MLE parameters is again $\bigO(\abs{\p} \!\cdot\! \abs{\data})$.

\begin{algorithm}[tb]
   \caption{PC Entropy regularization}
   \label{alg:ent-reg}
   \begin{algorithmic}[1]
   \STATE {\bfseries Input:} A deterministic PC $\p$; flow $\flow_{n,c}(\data)$ for every edge $(n,c)$ in $\p$; hyperparameter $\tau$.
   \STATE {\bfseries Output:} A set of log-parameters, $\{\varphi_{n,c} : (n,c) \in \p\}$, which are the solution of \cref{eq:det-pc-ent-reg-mle}.
   \STATE $\forall n, \; \mathtt{node\_prob}[n] \leftarrow 0; \quad \mathtt{node\_prob}[n_r] \leftarrow 1$ $\;$ \slash\slash $n_r$ is the root node of $\p$
   
   \WHILE{\tikzmarknode{c0}{} not converge}
   
   \STATE $\forall n, \; \mathtt{entropy}[n] \leftarrow \text{The entropy of the sub-PC rooted at}~n$ (see \cref{alg:pc-ent} in \cref{sec:useful-lems})
   
   \ForEach
   \FOR{\tikzmarknode{c1}{} sum unit $n$ traversed in preorder (parent before children)}
   \ForOnly
   \STATE $d_i \leftarrow \flow_{n,c_i}(\data) / \abs{\data}; \quad b = \tau \cdot \mathtt{node\_prob}[n]$ $\;$ \slash\slash $\{c_i\}_{i=1}^{\ch(n)}$ is the set of children of $n$
   \STATE{Solve for $\{\varphi_{n,c_i}\}_{i=1}^{\abs{\ch(n)}}$ in the following set of equations ($y$ is a variable):
    \begin{align}
        \begin{cases}
            d_i e^{-\varphi_{n,c_i}} - b \cdot \varphi_{n,c_i} + b \cdot \mathtt{entropy}[c_i] = y \quad (\forall i \in \{1,\dots,\abs{\ch(n)}\}) \\
            \sum_{i=1}^{\abs{\ch(n)}} e^{\varphi_{n,c_i}} = 1
        \end{cases}
        \label{eq:ent-reg-eqs}
    \end{align}
   }
   \NoDo
   \FOR{\tikzmarknode{c2}{} $\text{each}~c\in\ch(n)~\text{and~each}~m\in\ch(c)$ \textbf{do} $\;$ \slash\slash Update \texttt{node\_prob} of grandchildren}
   \STATE $\mathtt{node\_prob}[m] \leftarrow \mathtt{node\_prob}[m] + e^{\varphi_{n,c}} \cdot \mathtt{node\_prob}[n]$
   \ENDFOR
   \ReDo
   \ENDFOR
   
   \ENDWHILE
\end{algorithmic}
\begin{tikzpicture}[overlay,remember picture]
    \draw[black,line width=0.6pt] ([xshift=-27pt,yshift=-4pt]c0.west) -- ([xshift=-27pt,yshift=-145pt]c0.west) -- ([xshift=-23pt,yshift=-145pt]c0.west);
   \draw[black,line width=0.6pt] ([xshift=-35pt,yshift=-4pt]c1.west) -- ([xshift=-35pt,yshift=-118pt]c1.west) -- ([xshift=-31pt,yshift=-118pt]c1.west);
   \draw[black,line width=0.6pt] ([xshift=-15pt,yshift=-4pt]c2.west) -- ([xshift=-15pt,yshift=-14pt]c2.west) -- ([xshift=-11pt,yshift=-14pt]c2.west);
\end{tikzpicture}
\vspace{-0.8em}
\end{algorithm}

\boldparagraph{Entropy regularization}
The hope for tractable PC entropy regularization comes from the fact that the entropy of a deterministic PC $\p$ can be exactly computed in $\bigO(\abs{\p})$ time \citep{vergari2021compositional,SEneurips20}. However, it is still unclear whether the entropy regularization objective $\LL_{\mathrm{ent}} (\params; \data, \tau)$ (\cref{eq:det-pc-ent-reg-mle}) can be tractably maximized. We answer this question with a mixture of positive and negative results: while the objective is multi-modal and the global optimal is hard to find, we propose an efficient algorithm that (i) guarantees convergence to a stationary point, and (ii) achieves high convergence rate in practice. We start with the negative result.


\begin{prop}
\label{prop:hardness-ent-reg}
There exists a deterministic PC $\p$, a hyperparameter $\tau$, and a dataset $\data$ such that $\LL_{\mathrm{ent}} (\params; \data, \tau)$ (\cref{eq:det-pc-ent-reg-mle}) is non-concave and has multiple local maximas.
\end{prop}

Proof is given in \cref{sec:proof-hardness-ent-reg}. Although global optimal solutions are generally infeasible, we propose an efficient algorithm that guarantees to find a stationary point of $\LL_{\mathrm{ent}} (\params; \data, \tau)$. Specifically, \cref{alg:ent-reg} takes as input a deterministic PC $\p$ and all its edge flows w.r.t. $\data$, and returns a set of learned log-parameters that correspond to a stationary point of the objective.\footnote{We compute parameters in the logarithm space for numerical stability.} In its main loop (lines 4-10), the algorithm alternates between two procedures: (i) compute the entropy of the distribution encoded by every node w.r.t. the current parameters (line 5),\footnote{This can be done by \cref{alg:pc-ent} shown in \cref{sec:useful-lems}. \cref{lem:det-pc-ent-decom} proves that \cref{alg:pc-ent} takes $\bigO(\abs{\p})$ time.} and (ii) update PC parameters with regard to the computed entropies (lines 6-10). Specifically, in the parameter update phase (\ie the second phase), the algorithm traverses every sum unit $n$ in preorder and updates its child parameters by maximizing the entropy regularization objective ($\LL_{\mathrm{ent}} (\params; \data, \tau)$) with all other parameters fixed. This is done by solving the set of equations in \cref{eq:ent-reg-eqs} using Newton's method (lines 7-8).\footnote{Details for solving \cref{eq:ent-reg-eqs} is given in \cref{sec:solving-entreg-eqs}.} In addition to the child nodes' entropy computed in the first phase, \cref{eq:ent-reg-eqs} uses the top-down probability of every unit $n$ (i.e., $\mathtt{node\_prob}[n]$), which is progressively updated in lines 9-10.

\begin{thm}
\label{thm:ent-reg-alg-correctness}
\cref{alg:ent-reg} converges monotonically to a stationary point of $\LL_{\mathrm{ent}} (\params; \data, \tau)$ (\cref{eq:det-pc-ent-reg-mle}).
\end{thm}

\vspace{-0.6em}

\begin{proof}
The high-level idea of the proof is to show that the parameter update phase (lines 6-10) optimizes a concave surrogate objective of $\LL_{\mathrm{ent}} (\params; \data, \tau)$, which is determined by the entropies computed in line 5. Specifically, we show that whenever the surrogate objective is improved, $\LL_{\mathrm{ent}} (\params; \data, \tau)$ is also improved. Since the surrogate objective is concave, it can be easily optimized. Therefore, \cref{alg:ent-reg} converges to a stationary point of $\LL_{\mathrm{ent}} (\params; \data, \tau)$. The detailed proof is in \cref{sec:proof-ent-reg-alg}.
\end{proof}

\cref{alg:ent-reg} can be regarded as a EM-like algorithm, where the E-step is the entropy computation phase (line 5) and the M-step is the parameter update phase (lines 6-10). Specifically, the E-step constructs a concave surrogate of the true objective ($\LL_{\mathrm{ent}} (\params; \data, \tau)$), and the M-step updates all parameters by maximizing the concave surrogate function. Although \cref{thm:ent-reg-alg-correctness} provides no convergence rate analysis, the outer loop typically takes 3-5 iterations to converge in practice. Furthermore, \cref{eq:ent-reg-eqs} can be solved with high precision in a few ($<\!10$) iterations. Therefore, compared to the computation of all flows w.r.t. $\data$, which takes $\bigO(\abs{\p}\!\cdot\!\abs{\data})$ time, \cref{alg:ent-reg} takes a negligible $\bigO(\abs{\p})$ time.

In response to the motivation in \cref{sec:pc}, we show that both proposed methods can overcome the imbalanced regularization problem of Laplace smoothing. Again consider the example PC in \cref{fig:pseudocount-problems}(a), we conceptually demonstrate that both data softening and entropy regularization will not over-regularization $\theta_{11}$ compared to $\theta_{12}$. First, data softening essentially add no prior to the parameters, and only soften the evidences in the dataset. Therefore, it will not over-regularize children with small support sizes. Second, entropy regularization will add a much higher prior to $\theta_{12}$. Suppose $n\!=\!10$, consider maximizing \cref{eq:det-pc-ent-reg-mle} with an empty dataset (\ie we maximize $\entropy(\p_{n_1})$ directly), the optimal parameters would be $\theta_{11}\!\approx\!0.002$ and $\theta_{12}\!\approx\!0.998$. Therefore, entropy regularization will tend to add a higher prior to children with large support sizes. More fundamentally, the reason why both proposed approaches do not add biased priors to PCs is that they are designed to be model-agnostic, \ie their definitions as shown in \cref{sec:intuitive-ideas} are independent with the model they apply to.


\begin{figure}[t]
    \centering
    \includegraphics[width=\columnwidth]{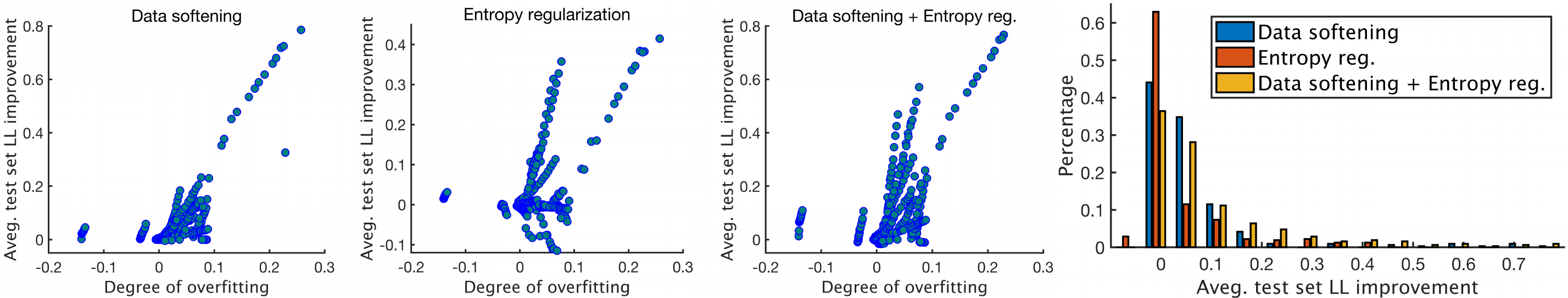}
    \vspace{-1.6em}
    \caption{Both data softening and entropy regularization effectively improve the test set log-likelihood (LL) across various datasets \citep{van2012markov} and PC structures \citep{dang2020strudel}. LL improvement (higher is better) represents the gain of test set LL compared to Laplace smoothing.}
    \label{fig:det-pc-exps}
\end{figure}

\boldparagraph{Empirical evaluation}
We empirically evaluate both proposed regularization methods on the twenty density estimation datasets \citep{van2012markov}. Since we are only concerned with parameter learning, we adopt PC structures (defined by its DAG) learned by Strudel \citep{dang2020strudel}. 16 PCs with different sizes were selected for each of the 20 datasets. For all experiments, we performed a hyperparameter search for all three regularization approaches (Laplace smoothing, data softening, and entropy regularization)\footnote{Specifically, $\alpha \in \{0.1,0.4,1.0,2.0,4.0,10.0\}$, $\beta \in \{0.9996,0.999,0.996\}$, $\tau \in \{0.001,0.01,0.1\}$.} using the validation set and report results on the test set. Please refer to \cref{sec:detail-det-pc-exps} for more details.

Results are summarized in \cref{fig:det-pc-exps}. First look at the scatter plots on the left. The x-axis represents the degree of overfitting, which is computed as follows: denote $\LL_{train}$ and $\LL_{val}$ as the average train and validation log-likelihood under the MLE estimation with Laplace smoothing ($\pseudocount\!=\!1.0$), the degree of overfitting is defined as $(\LL_{val} - \LL_{train}) / \LL_{val}$, which roughly captures how much the dataset/model pair suffers from overfitting. The y-axis represents the improvement on the average test set log-likelihood compared to Laplace smoothing. As demonstrated by the scatter plots, despite a few outliers, both proposed regularization methods steadily improve the test set LL over various datasets and PC structures, and the LL improvements are positively correlated with the degree of overfitting. Furthermore, as shown by the last scatter plot and the histogram plot, when combining data softening and entropy regularization, the LL improvement becomes much higher compared to using the two regularizers individually.

\subsection{Regularizing Non-Deterministic PCs}
\label{sec:new-reg-methods-nondet}

By viewing every non-deterministic PC as a deterministic PC with additional hidden variables (\cref{sec:pc-learning}), the regularization techniques developed in \cref{sec:new-reg-methods-det} can be directly adapted. Specifically, data softening can be regarded as injecting noise in both observed and hidden variables. Since the dataset provides no information about the hidden variables anyway, data softening essentially still ``perturbs'' the observed variables only. On the other hand, entropy regularization will have different behaviors when applied to non-deterministic PCs. Specifically, since it is coNP-hard to compute the entropy of a non-deterministic PC \citep{vergari2021compositional}, it is infeasible to optimize the entropy regularization objective $\LL_{\mathrm{ent}}(\params; \data, \tau)$ (\cref{eq:det-pc-ent-reg-mle}). However, we can still regularize the entropy of the distribution encoded by a non-deterministic PC over both of its observed and hidden variables, since explicitly representing the hidden variables renders the PC deterministic (\cref{sec:pc-learning}).

On the implementation side, data softening is performed by modifying the forward pass of the algorithm used to compute expected flows (\ie \cref{alg:nondet-forward-pass,alg:nondet-backward-pass} in the Appendix). Entropy regularization is again performed by \cref{alg:ent-reg} at the M-step of each min-batch/full-batch EM update, except that the input flows (\ie $\flow$) are replaced by the corresponding expected flows (\ie $\expflow$).






\begin{wrapfigure}{R}{0.4\textwidth}
\centering
\includegraphics[height=1.3in]{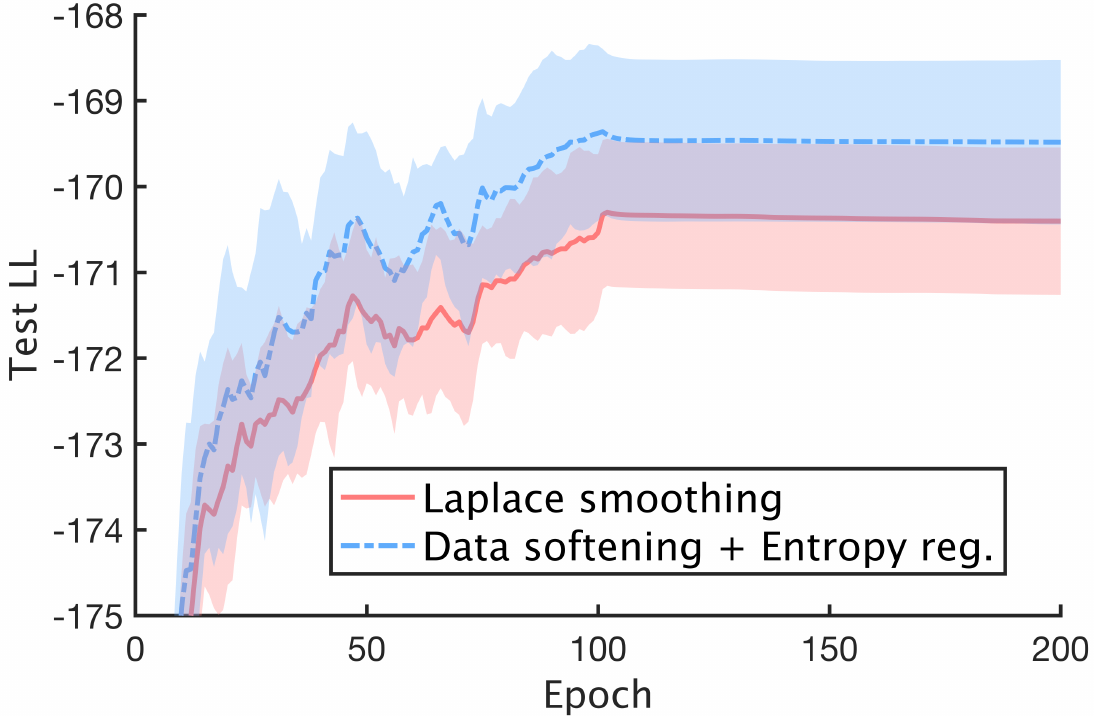}
\vspace{-0.6em}
\caption{Average ($\pm$std) test LL over 5 trials on the protein dataset.}
\label{fig:protein-exp}
\vspace{-1em}
\end{wrapfigure}

\boldparagraph{Empirical evaluation}
We use a simple yet effective PC structure, hidden Chow-Liu Tree (HCLT), as demonstrated in \cref{fig:nondet-pc3}. Specifically, on the left is a Bayesian network representation of a Chow-Liu Tree (CLT) \citep{chow1968approximating} over 5 variables. For any CLT over variables $\{X_i\}_{i=1}^{k}$, we can modify it as a HCLT through the following steps. First, we introduce a set of $k$ latent variables $\{Z_i\}_{i=1}^{k}$. Next, we replace all observed variables in the CLT with its corresponding latent variable (\ie $\forall i, X_i$ is replaced by $Z_i$). Finally, we add an edge from every latent variable to its corresponding observed variable (\ie $\forall i$, add an edge $Z_i \!\rightarrow\! X_i$). The HCLT structure is then compiled into a PC that encodes the same probability distribution. We used the hybrid mini-batch + full-batch EM as described in \cref{sec:pc-learning}. For all experiments, we trained the PCs with 100 mini-batch EM epochs and 100 full-batch EM epochs. Please refer to \cref{sec:nondet-exps-details} for hyperparameters related to the HCLT structure. Similar to \cref{sec:new-reg-methods-det}, we perform hyperparameter search for all methods using the validation set, and report results on the test set.

We first examine the performance on a protein sequence dataset \citep{russ2020evolution} that suffers from severe overfitting. Specifically, the training LL is typically above $-100$ while the validation and test set LL are around $-170$. \cref{fig:protein-exp} shows the test LL for Laplace smoothing and the hybrid regularization approach as training progresses. With the help of data softening and entropy regularization, we were able to obtain consistently higher test set LL. Next, we compare our HCLT model (with regularization) with the state-of-the-art PSDD (Strudel \citep{dang2020strudel} and LearnPSDD \citep{liang2017learning}) and SPN (EinSumNet \citep{peharz2020einsum}, LearnSPN \citep{gens2013learning}, ID-SPN \citep{rooshenas2014learning}, and RAT-SPN \citep{peharz2020random}) learning algorithms. With proper regularization, HCLT out-performed all baselines in 10 out of 20 datasets. Comparing with individual baselines, HCLT out-performs both PSDD learners on all datasets; HCLT achieved higher log-likelihood on 18, 19, 10, and 17 datasets compared to EinSumNet, LearnSPN, ID-SPN, and RAT-SPN, respectively.

\newcolumntype{R}{>{}r<{}}
\newcolumntype{L}{>{}l<{}}
\newcolumntype{M}{R@{}L}

\begin{table}[t]
    \caption{Test set log-likelihood in 20 density estimation benchmarks. We compare our method (HCLT) with the best performance (Best PSDD) over 2 deterministic PC learner: Strudel \citep{dang2020strudel} and LearnPSDD \citep{liang2017learning} as well as the best performance (Best SPN) over 4 SPN learning algorithms: EinSumNet \citep{peharz2020einsum}, LearnSPN \citep{gens2013learning}, ID-SPN \citep{rooshenas2014learning}, and RAT-SPN \citep{peharz2020random}. With the help of data softening and entropy regularization ($\pseudocount\!=\!0.1$, $\beta\!=\!0.002$, and $\tau\!=\!0.001$), HCLT achieved the best performance over 10 out of 20 datasets. All experiments for HCLT were repeated $5$ times, and the average and standard deviation are reported.
    }
    \label{tab:20datasets}
    \centering
    \scalebox{0.88}{
    {\renewcommand{\arraystretch}{0.8}
    \begin{tabular}{lMcc|lMcc}
        \toprule
        Dataset & \multicolumn{2}{c}{HCLT} & Best PSDD & Best SPN & Dataset & \multicolumn{2}{c}{HCLT} & Best PSDD & Best SPN \\
        \midrule
        accidents & \textbf{-26.74}&$\pm$0.03 & -28.29 & -26.98 & jester & \textbf{-52.46}&$\pm$0.01 & -54.63 & -52.56 \\
        ad & \textbf{-16.07}&$\pm$0.06 & -16.52 & -19.00 & kdd & -2.18&$\pm$0.00 & -2.17 & \textbf{-2.12} \\
        baudio & \textbf{-39.77}&$\pm$0.01 & -41.51 & -39.79 & kosarek & -10.66&$\pm$0.01 & -10.98 & \textbf{-10.60} \\
        bbc & -251.04&$\pm$1.19 & -258.96 & \textbf{-248.33} & msnbc & -6.05&$\pm$0.01 & -6.04 & \textbf{-6.03} \\
        bnetflix & \textbf{-56.27}&$\pm$0.01 & -58.53 & -56.36 & msweb & -9.98&$\pm$0.05 & -9.93 & \textbf{-9.73} \\
        book & \textbf{-33.83}&$\pm$0.01 & -35.77 & -34.14 & nltcs & \textbf{-5.99}&$\pm$0.01 & -6.03 & -6.01 \\
        c20ng & -153.40&$\pm$3.83 & -160.43 & \textbf{-151.47} & plants & -14.26&$\pm$0.16 & -13.49 & \textbf{-12.54} \\
        cr52 & -86.26&$\pm$3.67 & -92.38 & \textbf{-83.35} & pumbs* & -23.64&$\pm$0.25 & -25.28 & \textbf{-22.40} \\
        cwebkb & -152.77&$\pm$1.07 & -160.5 & \textbf{-151.84} & tmovie & \textbf{-50.81}&$\pm$0.12 & -55.41 & -51.51 \\
        dna & \textbf{-79.05}&$\pm$0.17 & -82.03 & -81.21 & tretail & \textbf{-10.84}&$\pm$0.01 & -10.90 & -10.85 \\
        \bottomrule
    \end{tabular}}}
\end{table}

\section{Conclusions}

This paper proposes two model-agnostic distribution regularization techniques: data softening and entropy regularization. While both methods are infeasible for many machine learning models, we theoretically show that they can be efficiently implemented when applied to probabilistic circuits. On the empirical side, we show that both proposed regularizers consistently improve the generalization performance over a wide variety of PC structures and datasets.

\bibliography{refs}

\begin{thebibliography}{48}
\providecommand{\natexlab}[1]{#1}
\providecommand{\url}[1]{\texttt{#1}}
\expandafter\ifx\csname urlstyle\endcsname\relax
  \providecommand{\doi}[1]{doi: #1}\else
  \providecommand{\doi}{doi: \begingroup \urlstyle{rm}\Url}\fi

\bibitem[Allenby \& Rossi(2006)Allenby and Rossi]{allenby2006hierarchical}
Allenby, G.~M. and Rossi, P.~E.
\newblock Hierarchical bayes models.
\newblock \emph{The handbook of marketing research: Uses, misuses, and future
  advances}, pp.\  418--440, 2006.

\bibitem[Chan \& Darwiche(2005)Chan and Darwiche]{chan2005revision}
Chan, H. and Darwiche, A.
\newblock On the revision of probabilistic beliefs using uncertain evidence.
\newblock \emph{Artificial Intelligence}, 163\penalty0 (1):\penalty0 67--90,
  2005.

\bibitem[Choi \& Darwiche(2017)Choi and Darwiche]{choi2017relaxing}
Choi, A. and Darwiche, A.
\newblock On relaxing determinism in arithmetic circuits.
\newblock In \emph{International Conference on Machine Learning}, pp.\
  825--833. PMLR, 2017.

\bibitem[Choi et~al.(2020{\natexlab{a}})Choi, Dang, and Broeck]{choi2020group}
Choi, Y., Dang, M., and Broeck, G. V.~d.
\newblock Group fairness by probabilistic modeling with latent fair decisions.
\newblock \emph{arXiv preprint arXiv:2009.09031}, 2020{\natexlab{a}}.

\bibitem[Choi et~al.(2020{\natexlab{b}})Choi, Farnadi, Babaki, and Van~den
  Broeck]{choi2020learning}
Choi, Y., Farnadi, G., Babaki, B., and Van~den Broeck, G.
\newblock Learning fair naive bayes classifiers by discovering and eliminating
  discrimination patterns.
\newblock In \emph{Proceedings of the AAAI Conference on Artificial
  Intelligence}, volume~34, pp.\  10077--10084, 2020{\natexlab{b}}.

\bibitem[Choi et~al.(2020{\natexlab{c}})Choi, Vergari, and Van~den
  Broeck]{choiprobabilistic}
Choi, Y., Vergari, A., and Van~den Broeck, G.
\newblock Probabilistic circuits: A unifying framework for tractable
  probabilistic models.
\newblock \emph{preprint}, 2020{\natexlab{c}}.

\bibitem[Chow \& Liu(1968)Chow and Liu]{chow1968approximating}
Chow, C. and Liu, C.
\newblock Approximating discrete probability distributions with dependence
  trees.
\newblock \emph{IEEE transactions on Information Theory}, 14\penalty0
  (3):\penalty0 462--467, 1968.

\bibitem[Dang et~al.(2020)Dang, Vergari, and Broeck]{dang2020strudel}
Dang, M., Vergari, A., and Broeck, G. V.~d.
\newblock Strudel: Learning structured-decomposable probabilistic circuits.
\newblock \emph{arXiv preprint arXiv:2007.09331}, 2020.

\bibitem[Dang et~al.(2021)Dang, Khosravi, Liang, Vergari, and Van~den
  Broeck]{dang2021juice}
Dang, M., Khosravi, P., Liang, Y., Vergari, A., and Van~den Broeck, G.
\newblock Juice: A julia package for logic and probabilistic circuits.
\newblock In \emph{Proceedings of the 35th AAAI Conference on Artificial
  Intelligence (Demo Track)}, 2021.

\bibitem[Darwiche(2003)]{darwiche2003differential}
Darwiche, A.
\newblock A differential approach to inference in bayesian networks.
\newblock \emph{Journal of the ACM (JACM)}, 50\penalty0 (3):\penalty0 280--305,
  2003.

\bibitem[Darwiche(2009)]{darwiche2009modeling}
Darwiche, A.
\newblock \emph{Modeling and reasoning with Bayesian networks}.
\newblock Cambridge university press, 2009.

\bibitem[Dempster et~al.(1977)Dempster, Laird, and Rubin]{dempster1977maximum}
Dempster, A.~P., Laird, N.~M., and Rubin, D.~B.
\newblock Maximum likelihood from incomplete data via the em algorithm.
\newblock \emph{Journal of the Royal Statistical Society: Series B
  (Methodological)}, 39\penalty0 (1):\penalty0 1--22, 1977.

\bibitem[Feng et~al.(2017)Feng, Wang, and Liu]{feng2017learning}
Feng, Y., Wang, D., and Liu, Q.
\newblock Learning to draw samples with amortized stein variational gradient
  descent.
\newblock \emph{arXiv preprint arXiv:1707.06626}, 2017.

\bibitem[Gal \& Ghahramani(2016)Gal and Ghahramani]{gal2016dropout}
Gal, Y. and Ghahramani, Z.
\newblock Dropout as a bayesian approximation: Representing model uncertainty
  in deep learning.
\newblock In \emph{international conference on machine learning}, pp.\
  1050--1059. PMLR, 2016.

\bibitem[Gens \& Pedro(2013)Gens and Pedro]{gens2013learning}
Gens, R. and Pedro, D.
\newblock Learning the structure of sum-product networks.
\newblock In \emph{International conference on machine learning}, pp.\
  873--880. PMLR, 2013.

\bibitem[Goan \& Fookes(2020)Goan and Fookes]{goan2020bayesian}
Goan, E. and Fookes, C.
\newblock Bayesian neural networks: An introduction and survey.
\newblock In \emph{Case Studies in Applied Bayesian Data Science}, pp.\
  45--87. Springer, 2020.

\bibitem[Goodfellow et~al.(2014)Goodfellow, Pouget-Abadie, Mirza, Xu,
  Warde-Farley, Ozair, Courville, and Bengio]{goodfellow2014generative}
Goodfellow, I.~J., Pouget-Abadie, J., Mirza, M., Xu, B., Warde-Farley, D.,
  Ozair, S., Courville, A.~C., and Bengio, Y.
\newblock Generative adversarial nets.
\newblock In \emph{NIPS}, 2014.

\bibitem[Grandvalet \& Bengio(2006)Grandvalet and
  Bengio]{grandvalet2006entropy}
Grandvalet, Y. and Bengio, Y.
\newblock Entropy regularization., 2006.

\bibitem[Heckerman(2008)]{heckerman2008tutorial}
Heckerman, D.
\newblock A tutorial on learning with bayesian networks.
\newblock \emph{Innovations in Bayesian networks}, pp.\  33--82, 2008.

\bibitem[Ioffe \& Szegedy(2015)Ioffe and Szegedy]{ioffe2015batch}
Ioffe, S. and Szegedy, C.
\newblock Batch normalization: Accelerating deep network training by reducing
  internal covariate shift.
\newblock In \emph{International conference on machine learning}, pp.\
  448--456. PMLR, 2015.

\bibitem[Jeffrey(1990)]{jeffrey1990logic}
Jeffrey, R.~C.
\newblock \emph{The logic of decision}.
\newblock University of Chicago press, 1990.

\bibitem[Khosravi et~al.(2019)Khosravi, Choi, Liang, Vergari, and Van~den
  Broeck]{KhosraviNeurIPS19}
Khosravi, P., Choi, Y., Liang, Y., Vergari, A., and Van~den Broeck, G.
\newblock On tractable computation of expected predictions.
\newblock In \emph{Advances in Neural Information Processing Systems 32
  (NeurIPS)}, dec 2019.

\bibitem[Kingma \& Welling(2013)Kingma and Welling]{kingma2013auto}
Kingma, D.~P. and Welling, M.
\newblock Auto-encoding variational bayes.
\newblock \emph{arXiv preprint arXiv:1312.6114}, 2013.

\bibitem[Kisa et~al.(2014)Kisa, Van~den Broeck, Choi, and
  Darwiche]{kisa2014probabilistic}
Kisa, D., Van~den Broeck, G., Choi, A., and Darwiche, A.
\newblock Probabilistic sentential decision diagrams.
\newblock In \emph{Proceedings of the 14th international conference on
  principles of knowledge representation and reasoning (KR)}, pp.\  1--10,
  2014.

\bibitem[Liang \& Van~den Broeck(2017)Liang and Van~den Broeck]{LiangXAI17}
Liang, Y. and Van~den Broeck, G.
\newblock Towards compact interpretable models: Shrinking of learned
  probabilistic sentential decision diagrams.
\newblock In \emph{IJCAI 2017 Workshop on Explainable Artificial Intelligence
  (XAI)}, August 2017.
\newblock URL \url{http://starai.cs.ucla.edu/papers/LiangXAI17.pdf}.

\bibitem[Liang et~al.(2017)Liang, Bekker, and Van~den
  Broeck]{liang2017learning}
Liang, Y., Bekker, J., and Van~den Broeck, G.
\newblock Learning the structure of probabilistic sentential decision diagrams.
\newblock In \emph{Proceedings of the 33rd Conference on Uncertainty in
  Artificial Intelligence (UAI)}, 2017.

\bibitem[Mei et~al.(2018)Mei, Jiang, and Tu]{mei2018maximum}
Mei, J., Jiang, Y., and Tu, K.
\newblock Maximum a posteriori inference in sum-product networks.
\newblock In \emph{Proceedings of the AAAI Conference on Artificial
  Intelligence}, volume~32, 2018.

\bibitem[Molina et~al.(2019)Molina, Vergari, Stelzner, Peharz, Subramani,
  Di~Mauro, Poupart, and Kersting]{molina2019spflow}
Molina, A., Vergari, A., Stelzner, K., Peharz, R., Subramani, P., Di~Mauro, N.,
  Poupart, P., and Kersting, K.
\newblock Spflow: An easy and extensible library for deep probabilistic
  learning using sum-product networks.
\newblock \emph{arXiv preprint arXiv:1901.03704}, 2019.

\bibitem[Pan et~al.(2006)Pan, Peng, and Ding]{pan2006belief}
Pan, R., Peng, Y., and Ding, Z.
\newblock Belief update in bayesian networks using uncertain evidence.
\newblock In \emph{2006 18th IEEE International Conference on Tools with
  Artificial Intelligence (ICTAI'06)}, pp.\  441--444. IEEE, 2006.

\bibitem[Pearl(2014)]{pearl2014probabilistic}
Pearl, J.
\newblock \emph{Probabilistic reasoning in intelligent systems: networks of
  plausible inference}.
\newblock Elsevier, 2014.

\bibitem[Peharz et~al.(2014)Peharz, Gens, and Domingos]{peharz2014learning}
Peharz, R., Gens, R., and Domingos, P.
\newblock Learning selective sum-product networks.
\newblock In \emph{LTPM workshop}, volume~32, 2014.

\bibitem[Peharz et~al.(2016)Peharz, Gens, Pernkopf, and
  Domingos]{peharz2016latent}
Peharz, R., Gens, R., Pernkopf, F., and Domingos, P.
\newblock On the latent variable interpretation in sum-product networks.
\newblock \emph{IEEE transactions on pattern analysis and machine
  intelligence}, 39\penalty0 (10):\penalty0 2030--2044, 2016.

\bibitem[Peharz et~al.(2020{\natexlab{a}})Peharz, Lang, Vergari, Stelzner,
  Molina, Trapp, Van~den Broeck, Kersting, and Ghahramani]{peharz2020einsum}
Peharz, R., Lang, S., Vergari, A., Stelzner, K., Molina, A., Trapp, M., Van~den
  Broeck, G., Kersting, K., and Ghahramani, Z.
\newblock Einsum networks: Fast and scalable learning of tractable
  probabilistic circuits.
\newblock In \emph{International Conference on Machine Learning}, pp.\
  7563--7574. PMLR, 2020{\natexlab{a}}.

\bibitem[Peharz et~al.(2020{\natexlab{b}})Peharz, Vergari, Stelzner, Molina,
  Shao, Trapp, Kersting, and Ghahramani]{peharz2020random}
Peharz, R., Vergari, A., Stelzner, K., Molina, A., Shao, X., Trapp, M.,
  Kersting, K., and Ghahramani, Z.
\newblock Random sum-product networks: A simple and effective approach to
  probabilistic deep learning.
\newblock In \emph{Uncertainty in Artificial Intelligence}, pp.\  334--344.
  PMLR, 2020{\natexlab{b}}.

\bibitem[Perez \& Wang(2017)Perez and Wang]{perez2017effectiveness}
Perez, L. and Wang, J.
\newblock The effectiveness of data augmentation in image classification using
  deep learning.
\newblock \emph{arXiv preprint arXiv:1712.04621}, 2017.

\bibitem[Poon \& Domingos(2011)Poon and Domingos]{poon2011sum}
Poon, H. and Domingos, P.
\newblock Sum-product networks: A new deep architecture.
\newblock In \emph{2011 IEEE International Conference on Computer Vision
  Workshops (ICCV Workshops)}, pp.\  689--690. IEEE, 2011.

\bibitem[Rooshenas \& Lowd(2014)Rooshenas and Lowd]{rooshenas2014learning}
Rooshenas, A. and Lowd, D.
\newblock Learning sum-product networks with direct and indirect variable
  interactions.
\newblock In \emph{International Conference on Machine Learning}, pp.\
  710--718. PMLR, 2014.

\bibitem[Russ et~al.(2020)Russ, Figliuzzi, Stocker, Barrat-Charlaix, Socolich,
  Kast, Hilvert, Monasson, Cocco, Weigt, et~al.]{russ2020evolution}
Russ, W.~P., Figliuzzi, M., Stocker, C., Barrat-Charlaix, P., Socolich, M.,
  Kast, P., Hilvert, D., Monasson, R., Cocco, S., Weigt, M., et~al.
\newblock An evolution-based model for designing chorismate mutase enzymes.
\newblock \emph{Science}, 369\penalty0 (6502):\penalty0 440--445, 2020.

\bibitem[Shen et~al.(2016)Shen, Choi, and Darwiche]{shen2016tractable}
Shen, Y., Choi, A., and Darwiche, A.
\newblock Tractable operations for arithmetic circuits of probabilistic models.
\newblock In \emph{Proceedings of the 30th International Conference on Neural
  Information Processing Systems}, pp.\  3943--3951. Citeseer, 2016.

\bibitem[Shih \& Ermon(2020)Shih and Ermon]{SEneurips20}
Shih, A. and Ermon, S.
\newblock Probabilistic circuits for variational inference in discrete
  graphical models.
\newblock In \emph{Advances in Neural Information Processing Systems 33
  (NeurIPS)}, december 2020.
\newblock URL
  \url{https://cs.stanford.edu/~andyshih/assets/pdf/SEneurips20.pdf}.

\bibitem[Srivastava et~al.(2014)Srivastava, Hinton, Krizhevsky, Sutskever, and
  Salakhutdinov]{srivastava2014dropout}
Srivastava, N., Hinton, G., Krizhevsky, A., Sutskever, I., and Salakhutdinov,
  R.
\newblock Dropout: a simple way to prevent neural networks from overfitting.
\newblock \emph{The journal of machine learning research}, 15\penalty0
  (1):\penalty0 1929--1958, 2014.

\bibitem[Szegedy et~al.(2016)Szegedy, Vanhoucke, Ioffe, Shlens, and
  Wojna]{szegedy2016rethinking}
Szegedy, C., Vanhoucke, V., Ioffe, S., Shlens, J., and Wojna, Z.
\newblock Rethinking the inception architecture for computer vision.
\newblock In \emph{Proceedings of the IEEE conference on computer vision and
  pattern recognition}, pp.\  2818--2826, 2016.

\bibitem[Van~den Broeck et~al.(2021)Van~den Broeck, Lykov, Schleich, and
  Suciu]{VdBAAAI21}
Van~den Broeck, G., Lykov, A., Schleich, M., and Suciu, D.
\newblock On the tractability of {SHAP} explanations.
\newblock In \emph{Proceedings of the 35th AAAI Conference on Artificial
  Intelligence}, Feb 2021.
\newblock URL \url{http://starai.cs.ucla.edu/papers/VdBAAAI21.pdf}.

\bibitem[Van~Haaren \& Davis(2012)Van~Haaren and Davis]{van2012markov}
Van~Haaren, J. and Davis, J.
\newblock Markov network structure learning: A randomized feature generation
  approach.
\newblock In \emph{Proceedings of the AAAI Conference on Artificial
  Intelligence}, volume~26, 2012.

\bibitem[Vergari et~al.(2015)Vergari, Di~Mauro, and
  Esposito]{vergari2015simplifying}
Vergari, A., Di~Mauro, N., and Esposito, F.
\newblock Simplifying, regularizing and strengthening sum-product network
  structure learning.
\newblock In \emph{Joint European Conference on Machine Learning and Knowledge
  Discovery in Databases}, pp.\  343--358. Springer, 2015.

\bibitem[Vergari et~al.(2021)Vergari, Choi, Liu, Teso, and Van~den
  Broeck]{vergari2021compositional}
Vergari, A., Choi, Y., Liu, A., Teso, S., and Van~den Broeck, G.
\newblock A compositional atlas of tractable circuit operations: From simple
  transformations to complex information-theoretic queries.
\newblock \emph{arXiv preprint arXiv:2102.06137}, 2021.

\bibitem[Vincent et~al.(2008)Vincent, Larochelle, Bengio, and
  Manzagol]{vincent2008extracting}
Vincent, P., Larochelle, H., Bengio, Y., and Manzagol, P.-A.
\newblock Extracting and composing robust features with denoising autoencoders.
\newblock In \emph{Proceedings of the 25th international conference on Machine
  learning}, pp.\  1096--1103, 2008.

\bibitem[Zhu et~al.(2017)Zhu, Park, Isola, and Efros]{zhu2017unpaired}
Zhu, J.-Y., Park, T., Isola, P., and Efros, A.~A.
\newblock Unpaired image-to-image translation using cycle-consistent
  adversarial networks.
\newblock In \emph{Proceedings of the IEEE international conference on computer
  vision}, pp.\  2223--2232, 2017.

\end{thebibliography}
\bibliographystyle{icml2021}

\clearpage
\appendix

\allowdisplaybreaks[4]

\section*{\centering \huge \bf Supplementary Material}

\section{Proofs}

This section provides the full proof of the theorems stated in the main paper.

\subsection{Proof of Theorem~\ref{thm:data-softening-alg}}
\label{sec:proof-data-soft-alg}

We break down the proof into two parts --- correctness of the forward pass (\cref{alg:forward-pass}) and correctness of the backward pass (\cref{alg:backward-pass}). As stated in the theorem, assume that we are given a deterministic PC $\p$, a boolean dataset $\data$ containing $N$ samples $\{\x^{(i)}\}_{i=1}^{N}$, and hyperparameter $\beta \in (0.5,1]$. Define $K$ as the number of variables in $\X$, \ie $\X = \{X_k\}_{k=1}^{K}$.

\boldparagraph{Correctness of the forward pass} 
We show that the value of each node $n$ w.r.t. sample $\x^{(i)}$ (by slightly abusing notation, denoted as $\mathtt{value}_{i}[n]$) computed by \cref{alg:forward-pass} (with the specific choice of $f_n(\x) = \beta \!\cdot\! \indicator{\x\in\supp(n)} + (1-\beta) \!\cdot\! \indicator{\x\not\in\supp(n)}$) is defined as
    \begin{align}
        \mathtt{value}_{i}[n] = \sum_{\x\in\val(\X)} \prod_{k=1}^{K} \Big ( \beta \!\cdot\! \indicator{x^{(i)}_{k} = x_{k}} + (1 \!-\! \beta) \!\cdot\! \indicator{x^{(i)}_{k} \neq x_{k}} \Big ) \cdot \indicator{\x\!\in\!\supp(n)},
        \label{eq:forward-pass-correctness}
    \end{align}
\noindent where $x_k$ denotes the $k$th feature of $\x$.
    
$\bullet$ Base case: input units. Suppose node $n$ is a literal w.r.t. variable $X_k$. That is, $\x\in\supp(n)$ iff $x_k=\lit(n)$, where $\lit(n)$ is either $\true$ or $\false$ defined by the PC. Denote $\neg \lit(n)$ as the negation of $\lit(n)$. $\forall i \in \{1,\dots,N\}$ we have 
    \begin{align*}
        \mathtt{value}_{i}[n] = & \beta \!\cdot\! \indicator{\x^{(i)}\in\supp(n)} + (1-\beta) \!\cdot\! \indicator{\x^{(i)}\not\in\supp(n)} \\
        = & \beta \!\cdot\! \indicator{x^{(i)}_{k} = \lit(n)} + (1-\beta) \!\cdot\! \indicator{x^{(i)}_{k} = \neg \lit(n)} \\
        \overset{(a)}{=} & \sum_{\x\in\{\x:\x\in\val(\X) \land x_k = \lit(n)\}} \prod_{l=1,l \neq k}^{K} \Big ( \beta \!\cdot\! \indicator{x^{(i)}_{l} = x_{l}} + (1 \!-\! \beta) \!\cdot\! \indicator{x^{(i)}_{l} \neq x_{l}} \Big ) \\
        & \qquad \qquad \cdot \Big ( \beta \!\cdot\! \indicator{x^{(i)}_{k} = \lit(n)} + (1-\beta) \!\cdot\! \indicator{x^{(i)}_{k} = \neg \lit(n)} \Big ) \\
        = & \sum_{\x\in\{\x:\x\in\val(\X) \land x_k = \lit(n)\}} \prod_{l=1,l \neq k}^{K} \Big ( \beta \!\cdot\! \indicator{x^{(i)}_{l} = x_{l}} + (1 \!-\! \beta) \!\cdot\! \indicator{x^{(i)}_{l} \neq x_{l}} \Big ) \\ 
        & \qquad \qquad \cdot \Big ( \beta \!\cdot\! \indicator{x^{(i)}_{k} = x_{k}} \!\cdot\! \indicator{x_{k} = \lit(n)} + (1-\beta) \!\cdot\! \indicator{x^{(i)}_{k} \neq x_{k}} \!\cdot\! \indicator{x_{k} = \lit(n)} \Big ) \\
        = & \sum_{\x\in\{\x:\x\in\val(\X) \land x_k = \lit(n)\}} \prod_{l=1,l \neq k}^{K} \Big ( \beta \!\cdot\! \indicator{x^{(i)}_{l} = x_{l}} + (1 \!-\! \beta) \!\cdot\! \indicator{x^{(i)}_{l} \neq x_{l}} \Big ) \\ 
        & \qquad \qquad \cdot \Big ( \beta \!\cdot\! \indicator{x^{(i)}_{k} = x_{k}} + (1-\beta) \!\cdot\! \indicator{x^{(i)}_{k} \neq x_{k}} \Big ) \!\cdot\! \indicator{x_{k} = \lit(n)} \\
        \overset{(b)}{=} & \sum_{\x\in\{\x:\x\in\val(\X)\}} \prod_{l=1}^{K} \Big ( \beta \!\cdot\! \indicator{x^{(i)}_{l} = x_{l}} + (1 \!-\! \beta) \!\cdot\! \indicator{x^{(i)}_{l} \neq x_{l}} \Big ) \!\cdot\! \indicator{x_{k} = \lit(n)} \\
        = & \sum_{\x\in\{\x:\x\in\val(\X)\}} \prod_{l=1}^{K} \Big ( \beta \!\cdot\! \indicator{x^{(i)}_{l} = x_{l}} + (1 \!-\! \beta) \!\cdot\! \indicator{x^{(i)}_{l} \neq x_{l}} \Big ) \!\cdot\! \indicator{\x\in\supp(n)},
    \end{align*}
\noindent where $(a)$ holds because the added term 
    \begin{align*}
        \sum_{\x\in\{\x:\x\in\val(\X) \land x_k = \lit(n)\}} \prod_{l=1,l \neq k}^{K} \Big ( \beta \!\cdot\! \indicator{x^{(i)}_{l} = x_{l}} + (1 \!-\! \beta) \!\cdot\! \indicator{x^{(i)}_{l} \neq x_{l}} \Big ) = 1;
    \end{align*}
\noindent the sum condition $\x_k=\lit(n)$ after $(b)$ can be lifted thanks to the indicator $\indicator{x_{k} = \lit(n)}$.

$\bullet$ Inductive case: product units. Suppose $n$ is a product unit with children $\{c_j\}_{j=1}^{\abs{\ch(n)}}$. Recall that the scope of the child $c_j$ is denoted as $\phi(c_j)$. Since the PC is decomposable, the contexts of different children are non-overlapping. Suppose the value of any child unit $c_j$ is defined according to \cref{eq:forward-pass-correctness}, \ie
    \begin{align*}
        \mathtt{value}_{i}[c_j] = \sum_{\x\in\val(\X)} \prod_{k=1}^{K} \Big ( \beta \!\cdot\! \indicator{x^{(i)}_{k} = x_{k}} + (1 \!-\! \beta) \!\cdot\! \indicator{x^{(i)}_{k} \neq x_{k}} \Big ) \cdot \indicator{\x\!\in\!\supp(c_j)}.
    \end{align*}
\noindent Denote $K_{c_j}$ as the set of index for the variables in $\phi(c_j)$. We have
    \begin{align*}
        \mathtt{value}_{i}[n] & \overset{(a)}{=} \prod_{j=1}^{\abs{\ch(n)}} \mathtt{value}_{i}[c_j] \\
        & = \prod_{j=1}^{\abs{\ch(n)}} \bigg \{ \sum_{\x\in\val(\X)} \prod_{k=1}^{K} \Big ( \beta \!\cdot\! \indicator{x^{(i)}_{k} = x_{k}} + (1 \!-\! \beta) \!\cdot\! \indicator{x^{(i)}_{k} \neq x_{k}} \Big ) \cdot \indicator{\x\!\in\!\supp(c_j)} \bigg \} \\
        & = \prod_{j=1}^{\abs{\ch(n)}} \bigg \{ \sum_{\x\in\val(\phi(c_j))} \prod_{k \in K_{c_j}} \Big ( \beta \!\cdot\! \indicator{x^{(i)}_{k} = x_{k}} + (1 \!-\! \beta) \!\cdot\! \indicator{x^{(i)}_{k} \neq x_{k}} \Big ) \cdot \indicator{\x\!\in\!\supp(c_j)} \bigg \} \\
        & \overset{(b)}{=} \sum_{\x\in\val(\bigcup_{j=1}^{\abs{\ch(n)}}\phi(c_j))} \prod_{k\in\bigcup_{j=1}^{\abs{\ch(n)}}K_{c_j}} \Big ( \beta \!\cdot\! \indicator{x^{(i)}_{k} = x_{k}} + (1 \!-\! \beta) \!\cdot\! \indicator{x^{(i)}_{k} \neq x_{k}} \Big ) \\
        & \qquad \qquad \qquad \qquad \qquad \quad \quad \; \cdot  \bigg ( \prod_{l=1}^{\abs{\ch(n)}} \indicator{\x\!\in\!\supp(c_l)} \bigg ) \\
        & \overset{(c)}{=} \sum_{\x\in\val(\bigcup_{j=1}^{\abs{\ch(n)}}\phi(c_j))} \prod_{k\in\bigcup_{j=1}^{\abs{\ch(n)}}K_{c_j}} \Big ( \beta \!\cdot\! \indicator{x^{(i)}_{k} = x_{k}} + (1 \!-\! \beta) \!\cdot\! \indicator{x^{(i)}_{k} \neq x_{k}} \Big ) \indicator{\x\!\in\!\supp(n)} \\
        & \overset{(d)}{=} \sum_{\x\in\val(\X)} \prod_{k=1}^{K} \Big ( \beta \!\cdot\! \indicator{x^{(i)}_{k} = x_{k}} + (1 \!-\! \beta) \!\cdot\! \indicator{x^{(i)}_{k} \neq x_{k}} \Big ) \indicator{\x\!\in\!\supp(n)},
    \end{align*}
\noindent where $(a)$ holds by line 6 of \cref{alg:forward-pass}; $(b)$ holds since $\forall c_i,c_j \in \ch(n) (c_i \neq c_j)$, we have $\phi(c_i) \cap \phi(c_j) = \emptyset$ and $K_{c_i} \cap K_{c_j} = \emptyset$ thanks to decomposability of the PC; $(c)$ is satisfied by the definition of product units: $\supp(n) = \bigcap_{c\in\ch(n)} \supp(c)$; $(d)$ holds since $\bigcup_{j=1}^{\abs{\ch(n)}}\phi(c_j)$ is a subset of $\X$.

$\bullet$ Inductive case: sum units. Suppose $n$ is a sum unit with children $\{c_j\}_{j=1}^{\abs{\ch(n)}}$. Suppose the value $\mathtt{value}_{i}[c_j]$ of any child unit $c_j$ is defined according to \cref{eq:forward-pass-correctness}, we have
    \begin{align*}
        \mathtt{value}_{i}[n] & \overset{(a)}{=} \sum_{j=1}^{\abs{\ch(n)}} \mathtt{value}_{i}[c_j] \\
        & = \sum_{j=1}^{\abs{\ch(n)}} \bigg \{ \sum_{\x\in\val(\X)} \prod_{k=1}^{K} \Big ( \beta \!\cdot\! \indicator{x^{(i)}_{k} = x_{k}} + (1 \!-\! \beta) \!\cdot\! \indicator{x^{(i)}_{k} \neq x_{k}} \Big ) \cdot \indicator{\x\!\in\!\supp(c_j)} \bigg \} \\
        & \overset{(b)}{=} \sum_{\x\in\val(\X)} \prod_{k=1}^{K} \Big ( \beta \!\cdot\! \indicator{x^{(i)}_{k} = x_{k}} + (1 \!-\! \beta) \!\cdot\! \indicator{x^{(i)}_{k} \neq x_{k}} \Big ) \cdot \Big ( \sum_{j=1}^{\abs{\ch(n)}} \indicator{\x\!\in\!\supp(c_j)} \Big ) \\
        & \overset{(c)}{=} \sum_{\x\in\val(\X)} \prod_{k=1}^{K} \Big ( \beta \!\cdot\! \indicator{x^{(i)}_{k} = x_{k}} + (1 \!-\! \beta) \!\cdot\! \indicator{x^{(i)}_{k} \neq x_{k}} \Big ) \cdot \indicator{\x\!\in\!\supp(n)},
    \end{align*}
\noindent where $(a)$ follows line 8 of \cref{alg:forward-pass}; $(b)$ holds because the sum unit $n$ is deterministic: $\forall c_i, c_j \in \ch(n) (c_i \neq c_j), \supp(c_i) \cap \supp(c_j) = \emptyset$; $(c)$ follows from the definition of sum units: $\supp(n) = \bigcup_{c\in\ch(n)} \supp(c)$.

We have shown that for any unit $n$, the value stored in $\mathtt{value}_{i}[n]$ follows the definition in \cref{eq:forward-pass-correctness}. We proceed to show the correctness of the backward pass.

\boldparagraph{Correctness of the backward pass}
Similar to the forward pass, we show that the context $\mathtt{context}_{i}[n]$ of each sum unit w.r.t. sample $\x^{(i)}$ computed by \cref{alg:backward-pass} is defined as
    \begin{align}
        \mathtt{context}_{i}[n] & = \sum_{\x\in\val(\X)} \prod_{k=1}^{K} \Big ( \beta \!\cdot\! \indicator{x^{(i)}_{k} = x_{k}} + (1 \!-\! \beta) \!\cdot\! \indicator{x^{(i)}_{k} \neq x_{k}} \Big ) \cdot \indicator{\x\!\in\!\context_n},
        \label{eq:backward-pass-correctness1}
    \end{align}
\noindent and the flow $\mathtt{flow}_{i}[n,c]$ of each edge $(n,c)$ s.t. $n$ is a sum unit is:
    \begin{align}
        \mathtt{flow}_{i}[n,c] & = \sum_{\x\in\val(\X)} \prod_{k=1}^{K} \Big ( \beta \!\cdot\! \indicator{x^{(i)}_{k} = x_{k}} + (1 \!-\! \beta) \!\cdot\! \indicator{x^{(i)}_{k} \neq x_{k}} \Big ) \cdot \indicator{\x\!\in\!\context_n \wedge \x\!\in\!\context_c}. 
        \label{eq:backward-pass-correctness2}
    \end{align}
    
$\bullet$ Base case: root unit $n_r$. Without loss of generality, we assume the root node represents a sum unit.\footnote{Note that if the root unit is not a sum, we can always add a sum unit as its parent and set the corresponding edge parameter to 1.} According to \cref{def:context}, the context of the root node $n_r$ equals its support, \ie $\context_{n_r} = \supp(n_r)$. Since in line 3 of \cref{alg:backward-pass}, the value $\mathtt{context}_{i}[n]$ is set to $\mathtt{value}_{i}[n]$, we know that
    \begin{align*}
        \mathtt{context}_{i}[n] & = \sum_{\x\in\val(\X)} \prod_{k=1}^{K} \Big ( \beta \!\cdot\! \indicator{x^{(i)}_{k} = x_{k}} + (1 \!-\! \beta) \!\cdot\! \indicator{x^{(i)}_{k} \neq x_{k}} \Big ) \cdot \indicator{\x\!\in\!\supp(n)} \\
        & = \sum_{\x\in\val(\X)} \prod_{k=1}^{K} \Big ( \beta \!\cdot\! \indicator{x^{(i)}_{k} = x_{k}} + (1 \!-\! \beta) \!\cdot\! \indicator{x^{(i)}_{k} \neq x_{k}} \Big ) \cdot \indicator{\x\!\in\!\context_{n}}.
    \end{align*}
    
$\bullet$ Inductive case: sum unit. Suppose $n$ is a sum unit with parent product units $\{m_j\}_{j=1}^{\abs{\pa(n)}}$. Denote the parent of product unit $m_i$ as $g_i$.\footnote{W.l.o.g. we assume all product unit only have one parent.} Suppose the contexts of $\{g_j\}_{j=1}^{\abs{\pa(n)}}$ satisfy \cref{eq:backward-pass-correctness1}. For ease of presentation, denote $H(\x,\x^{(i)},k) := \big ( \beta \!\cdot\! \indicator{x^{(i)}_{k} = x_{k}} + (1-\beta) \!\cdot\! \indicator{x^{(i)}_{k} \neq x_{k}} \big )$.
    \begin{align*}
        \mathtt{flow}_{i}[g_j,m_j] = & \frac{\mathtt{value}_{i}[m_j]}{\mathtt{value}_{i}[g_j]} \cdot \mathtt{context}_{i}[g_j] \\
        = & \frac{\sum\nolimits_{\x\in\val(\X)} \prod\nolimits_{k=1}^{K} H(\x,\x^{(i)},k) \cdot \indicator{\x\!\in\!\context_{g_j}}}{\sum\nolimits_{\x\in\val(\X)} \prod\nolimits_{k=1}^{K} H(\x,\x^{(i)},k) \cdot \indicator{\x\!\in\!\supp(g_j)}} \\
        & \qquad \qquad \qquad \qquad \cdot \sum_{\x\in\val(\X)} \prod_{k=1}^{K} H(\x,\x^{(i)},k) \cdot \indicator{\x\!\in\!\supp(m_j)} \numberthis\label{eq:thm1-proof-eq1}
    \end{align*}
Define $\context'_{g_j} := \bigcup_{c\in\pa(g_j)} \context_{c}$, \cref{def:context} suggests that $\context_{g_j} = \context'_{g_j} \cap \supp(g_j)$. Thus,
    \begin{align}
        & \indicator{\x\!\in\!\context_{g_j}} = \indicator{\x\!\in\!\context'_{g_j}} \cdot \indicator{\x\!\in\!\supp(g_j)}.
        \label{eq:thm1-proof-eq2}
    \end{align}
Consider conditioning $\supp(g_j)$ and $\context'_{g_j}$ on the variables $\phi(g_j)$ (\ie the variable scope of $g_j$). For any partial variable assignment $\e$ over $\phi(g_j)$, if $\e \in \supp(g_j)$, then $\e \in \context'_{g_j}$. Denote $K_{g_j}$ as the set of index for the variables in $\phi(g_j)$. We have
    \begin{align*}
        & \sum\limits_{\x\in\val(\X)} \prod\limits_{k=1}^{K} H(\x,\x^{(i)},k) \cdot \indicator{\x\!\in\!\context'_{g_j}} \cdot \indicator{\x\!\in\!\supp(g_j)} \\
        = & \bigg ( \sum\limits_{\x\in\val(\phi(g_j))} \prod\limits_{k \in K_{g_j}} H(\x,\x^{(i)},k) \cdot \indicator{\x\!\in\!\supp(g_j)} \bigg ) \\
        & \quad \cdot \bigg ( \sum\limits_{\x\in\val(\X\backslash\phi(g_j))} \prod\limits_{k \in \{1,\dots,K\} \backslash K_{g_j}} H(\x,\x^{(i)},k) \cdot \indicator{\x\!\in\!\context'_{g_j}} \bigg ) \numberthis\label{eq:thm1-proof-eq3}
    \end{align*}
Plug \cref{eq:thm1-proof-eq2,eq:thm1-proof-eq3} into \cref{eq:thm1-proof-eq1}, we have
    \begin{align*}
        \mathtt{flow}_{i}[g_j,m_j] & = \frac{\sum\nolimits_{\x\in\val(\phi(g_j))} \prod\nolimits_{k \in K_{g_j}} H(\x,\x^{(i)},k) \cdot \indicator{\x\!\in\!\supp(g_j)}}{\sum\nolimits_{\x\in\val(\X)} \prod\nolimits_{k=1}^{K} H(\x,\x^{(i)},k) \cdot \indicator{\x\!\in\!\supp(g_j)}} \\
        & \quad \cdot \bigg ( \sum\limits_{\x\in\val(\X\backslash\phi(g_j))} \prod\limits_{k \in \{1,\dots,K\} \backslash K_{g_j}} H(\x,\x^{(i)},k) \cdot \indicator{\x\!\in\!\context'_{g_j}} \bigg ) \\
        & \quad \cdot \bigg ( \sum_{\x\in\val(\X)} \prod_{k=1}^{K} H(\x,\x^{(i)},k) \cdot \indicator{\x\!\in\!\supp(m_j)} \bigg ) \\
        & = \bigg ( \sum\limits_{\x\in\val(\X\backslash\phi(g_j))} \prod\limits_{k \in \{1,\dots,K\} \backslash K_{g_j}} H(\x,\x^{(i)},k) \cdot \indicator{\x\!\in\!\context'_{g_j}} \bigg ) \\
        & \quad \cdot \bigg ( \sum_{\x\in\val(\phi(g_j))} \prod_{k \in K_{g_j}} H(\x,\x^{(i)},k) \cdot \indicator{\x\!\in\!\supp(m_j)} \bigg ) \numberthis\label{eq:thm1-proof-eq4}
    \end{align*}
Since $m_j$ is a child of $g_j$, the support of $m_j$ is a subset of $g_j$'s support: $\supp(m_j) \subseteq \supp(g_j)$. Therefore, for any partial variable assignment $\e$ over $\phi(g_j)$, if $\e \in \supp(m_j)$, then $\e \in \supp(g_j)$. Since $\{\e \mid \e\in\val(\phi(g_j)) \wedge \e\in\supp(g_j)\} \subseteq \{\e \mid \e\in\val(\phi(g_j)) \wedge \e\in\context'_{g_j}\}$, we conclude that for any partial variable assignment $\e$ over $\phi(g_j)$, if $\e \in \supp(m_j)$, then $\e \in \context'_{g_j}$. Therefore, the two product terms in \cref{eq:thm1-proof-eq4} can be joined with a Cartesian product:
    \begin{align}
        \mathtt{flow}_{i}[g_j,m_j] & = \sum_{\x\in\val(\X)} \prod_{k=1}^{K} H(\x,\x^{(i)},k) \cdot \indicator{\x\in\context'_{g_j} \cap \supp(m_j)}.
        \label{eq:thm1-proof-eq5}
    \end{align}
Note that $\context'_{g_j} \cap \supp(m_j) = \context'_{g_j} \cup \supp(g_j) \cap \supp(m_j) = \context_{g_j} \cap \supp(m_j)$. Since $\context_{m_j} = \context_{g_j} \cap \supp(m_j)$ (according to \cref{def:context}), we have 
    \begin{align*}
        \context'_{g_j} \cap \supp(m_j) & = \context_{g_j} \cap \supp(m_j) \\
        & = \context_{g_j} \cap \supp(g_j) \cap \supp(m_j) \\
        & = \context_{g_j} \cap \context_{m_j}.
    \end{align*}
Plug the above equation into \cref{eq:thm1-proof-eq5}, we have 
    \begin{align*}
        \mathtt{flow}_{i}[g_j,m_j] & = \sum_{\x\in\val(\X)} \prod_{k=1}^{K} H(\x,\x^{(i)},k) \cdot \indicator{\x\in\context_{g_j} \cap \context_{m_j}},
    \end{align*}
\noindent which is equivalent to \cref{eq:backward-pass-correctness1}.
    
We proceed to show that the context of unit $n$ follows \cref{eq:backward-pass-correctness2}. According to lines 6 and 7 of \cref{alg:backward-pass}, $\mathtt{context}_{i}[n]$ is computed as
    \begin{align*}
        \mathtt{context}_{i}[n] & = \sum_{j=1}^{\abs{\pa(n)}} \mathtt{flow}_{i}[g_j,m_j] \\
        & = \sum_{j=1}^{\abs{\pa(n)}} \sum_{\x\in\val(\X)} \prod_{k=1}^{K} H(\x,\x^{(i)},k) \cdot \indicator{\x\in\context_{g_j} \cap \context_{m_j}} \\
        & = \sum_{\x\in\val(\X)} \prod_{k=1}^{K} H(\x,\x^{(i)},k) \cdot \Big ( \sum_{j=1}^{\abs{\pa(n)}} \indicator{\x\in\context_{g_j} \cap \context_{m_j}} \Big ).
        \numberthis\label{eq:thm1-proof-eq6}
    \end{align*}
Next, we show that $\forall m_i, m_j \in \pa(n) (m_i \neq m_j), \context_{m_i} \cap \context_{m_j} = \emptyset$. We prove this claim using its contrapositive form. Suppose there exists $\x\in\val(\X)$ such that $\x\in\context_{m_i}$ and $\x\in\context_{m_j}$. According to the definition of context, if $\x\in\context_{m_i}$, then there must be a path between $m_i$ and the root node $n_r$ where all nodes in the path are ``activated'', \ie for any unit $c$ in the path, $\x\in\context_c$. Similarly, there much exists a path of ``activated'' units between $m_j$ and $n_r$. We note that the two paths must share a set of identical nodes since their terminal are both the root node $n_r$. Therefore, there must exist a sum unit $n'$ along the intersection of the two path where at least two of its children are activated, \ie $\exists c_1, c_2 \in \ch(n') (c_1 \neq c_2)$, such that $\x\in\context_{c_1}$ and $\x\in\context_{c_2}$. This contradicts the assumption that the PC is deterministic. Therefore, the claim at the beginning of this paragraph holds. Thus,
    \begin{align*}
        \sum_{j=1}^{\abs{\pa(n)}} \indicator{\x\in\context_{g_j} \cap \context_{m_j}} & \overset{(a)}{=} \sum_{j=1}^{\abs{\pa(n)}} \indicator{\x\in\context_{m_j}} \\
        & \overset{(b)}{=} \indicator{\x \in \bigcup\nolimits_{j=1}^{\abs{\pa(n)}} \context_{m_j}} \\
        & \overset{(c)}{=} \indicator{\x \in \bigcup\nolimits_{j=1}^{\abs{\pa(n)}} \context_{m_j} \cap \supp(n)} \\
        & \overset{(d)}{=} \indicator{\x \in \context_{n}},
    \end{align*}
\noindent where $(a)$ follows from $\context_{m_j} \subseteq \context_{g_j}$; $(b)$ holds because the statement made in the previous paragraph (\ie $\forall m_i, m_j \in \pa(n) (m_i \neq m_j), \context_{m_i} \cap \context_{m_j} = \emptyset$); $(c)$ holds since $\supp(m_j) \subseteq \supp(n)$ and $\context_{m_j} \subseteq \supp(m_j)$; $(d)$ directly applies the definition of context (\ie \cref{def:context}).

Plug in \cref{eq:thm1-proof-eq6}, we have
    \begin{align*}
        \mathtt{context}_{i}[n] & = \sum_{\x\in\val(\X)} \prod_{k=1}^{K} H(\x,\x^{(i)},k) \cdot \Big ( \sum_{j=1}^{\abs{\pa(n)}} \indicator{\x\in\context_{g_j} \cap \context_{m_j}} \Big ) \\
        & = \sum_{\x\in\val(\X)} \prod_{k=1}^{K} H(\x,\x^{(i)},k) \cdot \indicator{\x\in\context_{n}}.
    \end{align*}

\boldparagraph{Computing $\flow_{n,c} (\data_{\beta})$}
Finally, we can compute $\flow_{n,c} (\data_{\beta})$ from the flows (\ie $\mathtt{flow}_{i}[n,c]$) computed by \cref{alg:backward-pass}:
    \begin{align*}
        \flow_{n,c} (\data_{\beta}) & = \sum_{\x\in\val(\X)} \weight{\data_{\beta}}{\x} \cdot \indicator{\x\!\in\!\context_n \wedge \x\!\in\!\context_c} \\
        & = \sum_{\x\in\val(\X)} \underbrace{\sum_{i=1}^{N} \prod_{k=1}^{K} \Big ( \beta \!\cdot\! \indicator{x^{(i)}_{k} = x_{k}} + (1 \!-\! \beta) \!\cdot\! \indicator{x^{(i)}_{k} \neq x_{k}} \Big )}_{\weight{\data_{\beta}}{\x}} \cdot \indicator{\x\!\in\!\context_n \wedge \x\!\in\!\context_c} \\
        & = \sum_{i=1}^{N} \underbrace{\sum_{\x\in\val(\X)} \prod_{k=1}^{K} \Big ( \beta \!\cdot\! \indicator{x^{(i)}_{k} = x_{k}} + (1 \!-\! \beta) \!\cdot\! \indicator{x^{(i)}_{k} \neq x_{k}} \Big ) \cdot \indicator{\x\!\in\!\context_n \wedge \x\!\in\!\context_c}}_{\mathtt{flow}_{i}[n,c]} \\
        & = \sum_{i=1}^{N} \mathtt{flow}_{i}[n,c].
    \end{align*}
Finally, we note that \cref{alg:forward-pass,alg:backward-pass} both run in time $\bigO(\abs{\p}\!\cdot\!\abs{\data})$.

\subsection{Useful Lemmas}
\label{sec:useful-lems}

This section provides several useful lemmas that are later used in the proof of \cref{thm:ent-reg-alg-correctness}.

\begin{algorithm}[t]
\caption{PC entropy}
\label{alg:pc-ent}
{\fontsize{9}{9} \selectfont
\begin{algorithmic}[1]

\STATE {\bfseries Input:} A deterministic PC $\p$

\STATE {\bfseries Output:} $\mathtt{entropy}[n] := \entropy(\p_n)$ for every unit $n$

\ForEach
\FOR{\tikzmarknode{e0}{} $n$ traversed in postorder}

\STATE \textbf{if} $n$ \textbf{isa} input unit \textbf{then} $\mathtt{entropy}[n] = \entropy(\p_n)$ \slash\slash entropy of the input distribution
\STATE \textbf{elif} $n$ \textbf{isa} product unit \textbf{then} $\mathtt{entropy}[n] = \sum_{c\in\ch(n)} \mathtt{entropy}[c]$
\STATE{\tikzmarknode{e1}{}\textbf{else} \slash\slash $n$ is a sum unit \textbf{then}
    \begin{align*}
        \mathtt{entropy}[n] = - \sum_{c\in\ch(n)} \theta_{n,c} \log \theta_{n,c} + \sum_{c\in\ch(n)} \theta_{n,c} \cdot \mathtt{entropy}[c]
    \end{align*}
    \vspace{-1em}}

\ENDFOR
\end{algorithmic}
}
\begin{tikzpicture}[overlay,remember picture]
    \draw[black,line width=0.6pt] ([xshift=-28pt,yshift=-4pt]e0.west) -- ([xshift=-28pt,yshift=-65pt]e0.west) -- ([xshift=-24pt,yshift=-65pt]e0.west);
   \draw[black,line width=0.6pt] ([xshift=4pt,yshift=-4pt]e1.west) -- ([xshift=4pt,yshift=-34pt]e1.west) -- ([xshift=8pt,yshift=-34pt]e1.west);
\end{tikzpicture}
\vspace{-0.8em}
\end{algorithm}

\begin{lem}
\label{lem:det-pc-ent-decom}
Given a deterministic PC $\p$ whose root node is $n_r$, its entropy $\entropy(\p) \overset{def}{=} \entropy(\p_{n_r})$ can be decomposed recursively as follows:
    \begin{align*}
        \entropy(\p_n) = 
        \begin{cases}
            \sum_{c\in\ch(n)} \big ( -\theta_{n,c} \log \theta_{n,c} + \theta_{n,c} \cdot \entropy(\p_c) \big ) & \text{if}~n~\text{is a sum unit}, \\
            \sum_{c\in\ch(n)} \entropy(\p_c) & \text{if}~n~\text{is a product unit},
        \end{cases}
    \end{align*}
\noindent where the entropy of an input unit is defined by the entropy of the corresponding univariate distribution. Following this decomposition, we construct \cref{alg:pc-ent} that computes the entropy of every nodes in a deterministic PC in $\bigO(\abs{\p})$ time. 
\end{lem}

\begin{proof}
We show the correctness of the entropy decomposition over a sum unit and a product unit respectively.

$\bullet$ Sum units. If $n$ is a sum unit:
    \begin{align*}
        \entropy(\p_n) & = - \sum_{\x\in\val(\phi(n))} \Big ( \sum_{c\in\ch(n)} \theta_{n,c} \p_{c} (\x) \Big ) \log \Big ( \sum_{c\in\ch(n)} \theta_{n,c} \p_{c} (\x) \Big ) \\
        & = - \sum_{\x\in\val(\phi(n))} \Big ( \sum_{c\in\ch(n)} \theta_{n,c} \p_{c} (\x) \indicator{\x\in\supp(c)} \Big ) \log \Big ( \sum_{c\in\ch(n)} \theta_{n,c} \p_{c} (\x) \indicator{\x\in\supp(c)} \Big ) \\
        & \overset{(a)}{=} - \sum_{\x\in\val(\phi(n))} \sum_{c\in\ch(n)} \indicator{\x\in\supp(c)} \cdot \theta_{n,c} \cdot \p_{c} (\x) \cdot \Big ( \log \theta_{n,c} + \log \p_c (\x) \Big ) \\
        & = - \sum_{c\in\ch(n)} \theta_{n,c} \log \theta_{n,c} \underbrace{\Big ( \sum_{\x\in\val(\phi(n))} \indicator{\x\in\supp(c)} \p_c (\x) \Big )}_{=1} \\
        & \qquad \qquad \qquad + \sum_{c\in\ch(n)} \theta_{n,c} \underbrace{\Big ( - \sum_{\x\in\val(\phi(n))} \p_c(\x) \log \p_c(\x) \Big )}_{=\entropy(\p_c)} \\
        & = \sum_{c\in\ch(n)} \Big ( - \theta_{n,c} \log \theta_{n,c} + \theta_{n,c} \cdot \entropy(\p_c) \Big ), \numberthis\label{eq:prop2-proof-eq3}
    \end{align*}
\noindent where $(a)$ uses the assumption that the sum unit is deterministic, \ie $\forall c_1, c_2 \in \ch(n) \; (c_1 \neq c_2), \supp(c_1) \cap \supp(c_2) = \emptyset$.
    
$\bullet$ Product units. If $n$ is a product unit:
    \begin{align*}
        \entropy(\p_n) & = - \sum_{\x\in\val(\phi(n))} \Big ( \prod_{c\in\ch(n)} \p_{c} (\x) \Big ) \log \Big ( \prod_{c\in\ch(n)} \p_{c} (\x) \Big ) \\
        & = - \sum_{c\in\ch(n)} \Big ( \sum_{\x\in\val(\phi(c))} \p_c (\x) \log \p_c (\x) \Big ) \\
        & = \sum_{c\in\ch(n)} \entropy(\p_c).
        \numberthis\label{eq:prop2-proof-eq4}
    \end{align*}
\end{proof}

\begin{lem}
\label{lem:ent-non-concave}
The entropy of a deterministic PC $\p$ is neither convex nor concave w.r.t. its parameters.
\end{lem}

\begin{figure}
    \centering
    \includegraphics[width=0.5\columnwidth]{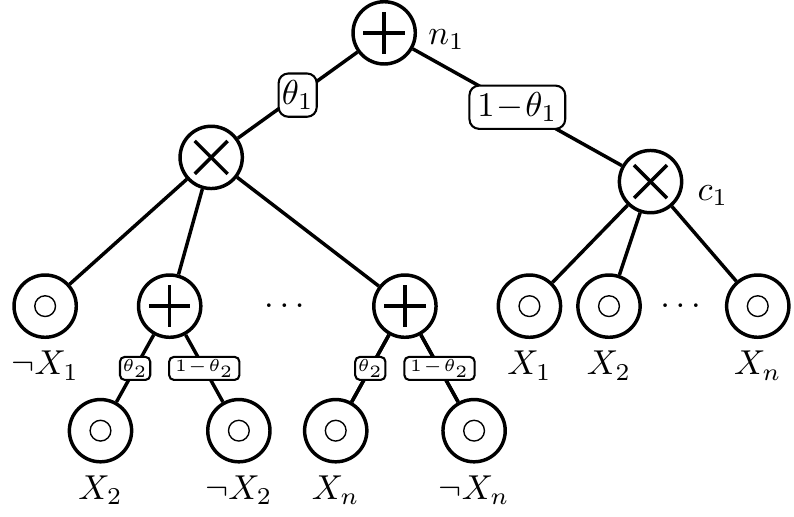}
    \caption{An example PC to show that PC entropy is neither convex nor concave.}
    \label{fig:example-pc-ent}
\end{figure}

\begin{proof}
Consider the example PC in \cref{fig:example-pc-ent}. Assume $n = 20$ and define parameters $\params_a = \{\theta_1 = 0.1, \theta_2 = 0.1\}$ and $\params_b = \{\theta_1 = 0.12, \theta_2 = 0.12\}$. Denote $\params_c = (\params_a + \params_b) / 2$, we have
    \begin{align*}
        2 \cdot \entropy(\p;\params_c) - \entropy(\p;\params_a) - \entropy(\p;\params_b) \approx -0.0047898 < 0.
    \end{align*}
Hence the entropy is not concave.

Define parameters $\params_d = \{\theta_1 = 0.4, \theta_2 = 0.8\}$ and $\params_e = \{\theta_1 = 0.42, \theta_2 = 0.82\}$. Denote $\params_f = (\params_d + \params_e) / 2$, we have
    \begin{align*}
        2 \cdot \entropy(\p;\params_f) - \entropy(\p;\params_d) - \entropy(\p;\params_e) \approx 0.0056294 > 0.
    \end{align*}
Hence the entropy is not convex.
\end{proof}

\begin{lem}
\label{lem:mle-concavity}
For any dataset $\data = \{\x^{(i)}\}_{i=1}^{N}$ and any deterministic PC $\p$ with parameters $\params$, the following formula is concave w.r.t. $\params$:
    \begin{align}
        \sum_{i = 1}^{N} \log \p (\x^{(i)}; \params).
        \label{eq:mle-concave}
    \end{align}
\end{lem}

\begin{proof}
For any input $\x$, $\log \p(\x; \params)$ can be decomposed over sum and product units:

$\bullet$ Sum units. Suppose $n$ is a sum unit, then 
    \begin{align*}
        \log \p_n (\x;\params) & = \log \Big ( \sum_{c\in\ch(n)} \theta_{n,c} \cdot \p_{c} (\x) \Big ) \\
        & = \log \Big ( \sum_{c\in\ch(n)} \theta_{n,c} \cdot \p_{c} (\x) \indicator{\x\in\supp(c)} \Big ) \\
        & = \sum_{c\in\ch(n)} \indicator{\x\in\supp(c)} \big ( \log \theta_{n,c} + \log \p_{c} (\x) \big ),
        \numberthis\label{eq:prop2-proof-eq1}
    \end{align*}
\noindent where the last equation holds because unit $n$ is deterministic: $\forall c_i, c_j \in \ch(n) (c_i \neq c_j), \supp(c_i) \cap \supp(c_j) = \emptyset$.

$\bullet$ Product units. Suppose $n$ is a product unit, then
    \begin{align}
        \log \p_n (\x;\params) = \log \bigg ( \prod_{c\in\ch(n)} \p_{c} (\x) \bigg ) = \sum_{c\in\ch(n)} \log \p_{c} (\x).
        \label{eq:prop2-proof-eq2}
    \end{align}
    
According to \cref{eq:prop2-proof-eq1,eq:prop2-proof-eq2}, for any $\x\in\val(\X)$, $\log \p(\x;\params)$ can be decomposed into the sum over a set of log-parameters (\eg $\log \theta_{n,c}$). Therefore, \cref{eq:mle-concave} is concave.
\end{proof}

\begin{lem}
\label{lem:entropy-decomposition}
Given a deterministic PC $\p$ with root node $n_r$, its entropy $\entropy(\p)$ can be decomposed as follows:
    \begin{align*}
        \entropy(\p_{n_r}) = - \sum_{(n,c)\in\mathtt{edges}(\p_{n_r})} \Pr_{n_r}(n) \cdot \theta_{n,c} \log \theta_{n,c},
    \end{align*}
\noindent where $\mathtt{edges}(\p)$ denotes all edges $(n,c)$ in the PC with sum unit $n$; $\Pr_{n_r}(n)$ is defined in \cref{eq:def-pr-n}.
\end{lem}

\begin{proof}
We prove the lemma by induction.

$\bullet$ Base case. Suppose $m$ is a sum unit such that all its decendents are either input units or product unit. By definition, we have $\Pr_{m}(m) = 1$, and $\mathtt{edges}(\p_m) = \{(m,c) \mid c\in\ch(m)\}$. Thus,
    \begin{align*}
        - \sum_{(n,c)\in\mathtt{edges}(\p_{m})} \Pr_{m}(n) \cdot \theta_{n,c} \log \theta_{n,c}= - \sum_{c\in\ch(m)} \theta_{m,c} \log \theta_{m,c} = \entropy(\p_m).
    \end{align*}
    
$\bullet$ Inductive case: product units. Suppose $m$ is a product unit such that for each of its children $c\in\ch(m)$, we have
    \begin{align*}
        \entropy(\p_{c}) = - \sum_{(n',c')\in\mathtt{edges}(\p_c)} \Pr_{c}(n') \cdot \theta_{n',c'} \log \theta_{n',c'}.
    \end{align*}
Then by \cref{lem:det-pc-ent-decom} we know that
    \begin{align*}
        \entropy(\p_{m}) & = \sum_{c\in\ch(m)} \entropy(\p_{m}) \\
        & = - \sum_{c\in\ch(m)} \sum_{(n',c')\in\mathtt{edges}(\p_c)} \Pr_{c}(n') \cdot \theta_{n',c'} \log \theta_{n',c'} \\
        & \overset{(a)}{=} - \sum_{c\in\ch(m)} \sum_{(n',c')\in\mathtt{edges}(\p_c)} \Pr_{m}(n') \cdot \theta_{n',c'} \log \theta_{n',c'} \\
        & \overset{(b)}{=} - \sum_{(n',c')\in\mathtt{edges}(\p_m)} \Pr_{m}(n') \cdot \theta_{n',c'} \log \theta_{n',c'},
    \end{align*}
\noindent where $(a)$ holds since for any sum unit $n'$, $\Pr_{c}(n') = \Pr_{m}(n')$, and $(b)$ follows from the fact that $\mathtt{edges}(\p_m) = \bigcup_{c\in\ch(m)} \mathtt{edges}(\p_c)$.

$\bullet$ Inductive case: sum units. Suppose $m$ is a sum unit such that for each of its children $c\in\ch(m)$, we have
    \begin{align*}
        \entropy(\p_{c}) = - \sum_{(n',c')\in\mathtt{edges}(\p_c)} \Pr_{c}(n') \cdot \theta_{n',c'} \log \theta_{n',c'}.
    \end{align*}
Then by \cref{lem:det-pc-ent-decom} we have
    \begin{align*}
        \entropy(\p_m) & = \sum_{c\in\ch(m)} \big ( -\theta_{n,c} \log \theta_{n,c} + \theta_{n,c} \cdot \entropy(\p_c) \big ) \\
        & = \sum_{c\in\ch(m)} - \theta_{n,c} \log \theta_{n,c} - \sum_{c\in\ch(n)} \sum_{(n',c')\in\mathtt{edges}(\p_c)} \underbrace{\theta_{m,c} \cdot \Pr_{c}(n')}_{\Pr_{m}(n')} \cdot \theta_{n',c'} \log \theta_{n',c'} \\
        & = \sum_{c\in\ch(m)} - \Pr_{m}(m) \theta_{n,c} \log \theta_{n,c} - \sum_{c\in\ch(n)} \sum_{(n',c')\in\mathtt{edges}(\p_c)} \Pr_{m}(n') \cdot \theta_{n',c'} \log \theta_{n',c'} \\
        & \overset{(a)}{=} - \sum_{(n',c')\in\mathtt{edges}(\p_m)} \Pr_{m}(n') \cdot \theta_{n',c'} \log \theta_{n',c'},
    \end{align*}
\noindent where $(a)$ holds because $\mathtt{edges}(\p_m) = \big (\bigcup_{c\in\ch(m)} \mathtt{edges}(\p_c) \big ) \; \bigcup \; \big ( \{(m,c) \mid c\in\ch(m)\} \big )$.
\end{proof}

\begin{lem}
\label{lem:ent-multimodal}
The entropy regularization objective in \cref{eq:det-pc-ent-reg-mle} w.r.t. a deterministic PC $\p$ and a dataset $\data$ could have multiple local maximas.
\end{lem}

\begin{figure}
    \centering
    \includegraphics[width=0.85\columnwidth]{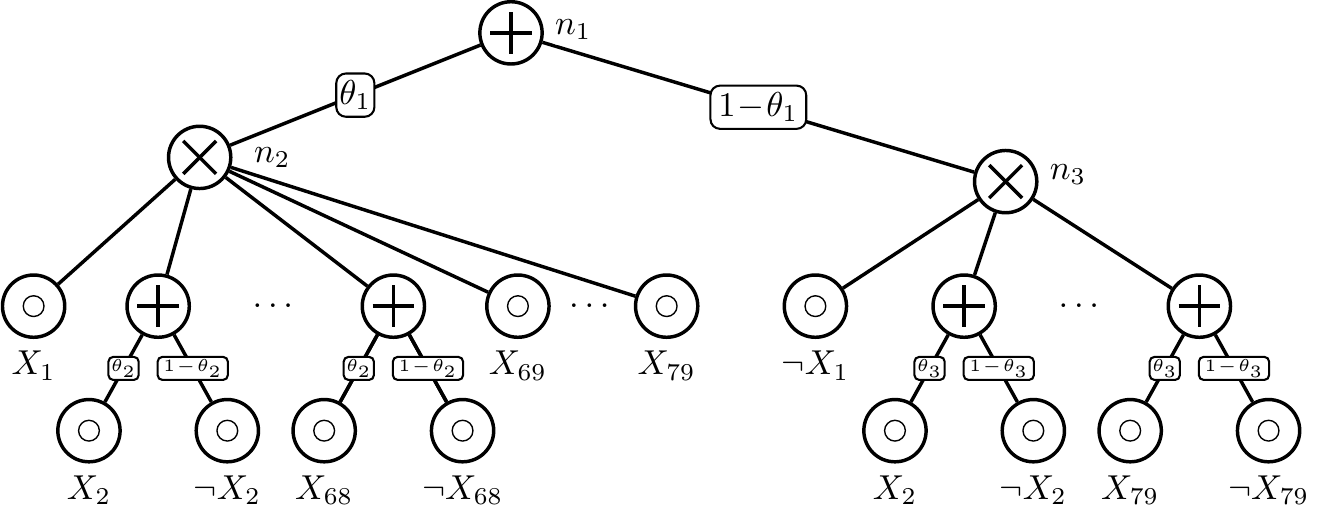}
    \caption{An example PC to show that \cref{eq:det-pc-ent-reg-mle} could have multiple stationary points.}
    \label{fig:example-pc-multimodal}
\end{figure}

\begin{proof}
Consider the deterministic PC $\p$ in \cref{fig:example-pc-multimodal} and dataset $\data$ with a single sample $\x = (\true, \dots, \true)$. The objective in \cref{eq:det-pc-ent-reg-mle} can be re-written as follows
    \begin{align*}
        \calL_{\mathrm{ent}} (\params; n_1, \tau) := \log \p_{n_1}(\x) + \tau \cdot \entropy(\p_{n_1}),
    \end{align*}
\noindent where $n_1$ is the root node of the PC as denoted in \cref{fig:example-pc-multimodal}. We further decompose $\calL_{\mathrm{ent}} (\params; n_1, \tau)$:
    \begin{align*}
        \calL_{\mathrm{ent}} (\params; n_1, \tau) = & \log \theta_1 + 67 \cdot \log \theta_2 + \tau \cdot \entropy(\p_{n_1}) \\
        = & \log \theta_1 + 67 \cdot \log \theta_2 + \tau \!\cdot\! \big ( - \theta_1 \log \theta_1 - (1\!-\!\theta_1) \log (1\!-\!\theta_1) \big ) \\
        & + \tau \cdot \theta_1 \cdot \entropy(\p_{n_2}) + \tau \cdot (1\!-\!\theta_1) \cdot \entropy(\p_{n_3}).
    \end{align*}
First, we observe that to maximize $\calL_{\mathrm{ent}} (\params; n_1, \tau)$, $\theta_3$ should always be $0.5$ since the only term that depends on $\theta_3$ is $(1\!-\!\theta_1) \cdot \entropy(\p_{n_3})$ and $1\!-\!\theta_1\!>\!0$. Therefore, we have
    \begin{align*}
        \entropy(\p_{n_3}) = 78 \cdot \log 2 \approx 54.065.
    \end{align*}
Next, for any fixed $\theta_1 \!\in\! (0,1]$, the objective $\calL_{\mathrm{ent}} (\params; n_1, \tau)$ is concave w.r.t. $\theta_2$:
    \begin{align}
        \calL_{\mathrm{ent}} (\params; n_1, \tau) = 67 \cdot \big ( \log \theta_2 - \tau \!\cdot\! \theta_1 \!\cdot\! (\theta_2 \log \theta_2 + (1\!-\!\theta_2) \log (1\!-\!\theta_2) ) \big ) + \mathrm{const},
        \label{eq:proof-lem5-eq-1}
    \end{align}
\noindent where the constant term does not depend on $\theta_2$. Therefore, for any $\theta_1$, we can uniquely compute the optimal value of $\theta_2$. We are left with determining the optimal value of $\theta_1$. Choose $\tau = 1.5$, the derivative of $\calL_{\mathrm{ent}} (\params; n_1, \tau)$ w.r.t. $\theta_1$ is (denote $\mathtt{ent}_0 := -(\theta_2 \log \theta_2 + (1\!-\!\theta_2) \log (1\!-\!\theta_2) )$)
    \begin{align*}
        g (\theta_1) := \frac{\partial \calL_{\mathrm{ent}} (\params; n_1, \tau)}{\partial \theta_1} & = \frac{1}{\theta_1} + 1.5 \cdot \Big ( \log (1\!-\!\theta_1) - \log \theta_1 + \entropy(\p_{n_2}) - \entropy(\p_{n_3}) \Big ) \\
        & = \frac{1}{\theta_1} + 1.5 \cdot \Big ( \log (1\!-\!\theta_1) - \log \theta_1 + 67 \cdot \mathtt{ent}_0 - \entropy(\p_{n_3}) \Big )
    \end{align*}
\noindent where $\entropy(\p_{n_3})$ can be viewed as a constant and $\mathtt{ent}_0$ depends on $\theta_1$. Specifically, for any $\theta_1$, we compute $\theta_2$ and hence $\mathtt{ent}_0$ by maximizing \cref{eq:proof-lem5-eq-1}. Putting everything together, we have
    \begin{align*}
        \begin{cases}
            g(0.02) \approx 1.772730 > 0, \\
            g(0.7) \approx -0.190743 < 0, \\
            g(0.9) \approx 2.055231 > 0, \\
            g(0.99) \approx -0.216938 < 0.
        \end{cases}
    \end{align*}
Since $g$ is continuous in range $(0,1]$, there exists a local maxima of $\theta_1$ between $0.02$ and $0.7$ as well as between $0.9$ and $0.99$. Therefore, the entropy regularization objective could have multiple local maximas.

\end{proof}

\subsection{Proof of Theorem~\ref{thm:hardness-soft-reg}}
\label{sec:proof-hardness-soft-reg}

This theorem is a direct corollary of Theorem 5 in \citep{VdBAAAI21}, which has the following statement:

Computing the expectation of a logistic regression model \wrt a uniform data distribution is \#-hard.

Note that with $\beta = 0.5$, the distribution $\data_{\beta}$ is essentially uniform, \cref{thm:hardness-soft-reg} follows directly from \citep{VdBAAAI21}.

\subsection{Proof of Theorem~\ref{thm:hardness-ent-reg}}
\label{sec:proof-sharpp-hardness-ent-reg}

This proof largely follows the proof of Theorem 5 in \citep{VdBAAAI21}. The proof is by reduction from \textsc{\#NUMPAR}, which is defined as follows. Given $n$ positive integers $k_1, \dots, k_n$, we want to count the number of subset $S \subseteq [n]$ that satisfies $\sum_{i \in S} k_i = \sum_{i \not\in S} k_i$. \textsc{\#NUMPAR} is known to be \#P-hard.

Fix an instance of \textsc{\#NUMPAR}, $k_1, \dots, k_n$, and assume w.l.o.g. that the sum of the numbers is even, \ie $\sum_i k_i = 2 c$ for some natural number $c$. Define $P := \{S \mid S \subseteq [n], \sum_{i \in S} k_i = c\}$. By definition $\abs{P}$ is the solution to the \textsc{\#NUMPAR} problem. Note that for each $S \in P$, its complement $\bar{S}$ should also be a member of $P$, and hence $\abs{P}$ is even.

Define a logistic regression model as $F(x_1, \dots, x_n) := \sigma(w_0 + \sum_{i=1}^{n} w_i \cdot x_i)$, where $\sigma$ is the sigmoid function. Define the normalized model of $F$ as $G(x_1, \dots, x_n) := F(x_1, \dots, x_n) / Z$, where $Z := \sum_{\x\in\val(X)} F(x_1, \dots, x_n)$. Denote the entropy of a normalized logistic regressor $G$ as $\entropy(G) := - \sum_{\x\in\val(\X)} G(\x) \log G(\x)$.

We now describe an algorithm that computes $\abs{P}$ using an oracle for $\entropy(G)$, where $G$ is a normalized logistic regression model. Denote $m$ as a large natural number to be chosen later, and define the following weights
    \begin{align*}
        w_0 := -\frac{m}{2} - mc, \quad w_i := m k_i (\forall i \in [n]).
    \end{align*}
Let $F$ be the logistic regressor corresponds to the above weights and $G$ the normalized model of $F$. We can represent $\entropy(G)$ as follows:
    \begin{align*}
        \entropy(G) = -\sum_{\x\in\val(\X)} \frac{F(\x)}{Z} \log \frac{F(\x)}{Z} & = -\sum_{\x\in\val(\X)} \Big ( \frac{F(\x) \log F(\x)}{Z} - \frac{F(\x)}{Z} \log Z \Big ) \\
        & = -\sum_{\x\in\val(\X)} \frac{F(\x) \log F(\x)}{Z} + \log Z.
    \end{align*}
For large enough $m$, $F(\x)$ will approach either $0$ or $1$. Therefore, the first term in the above equation will approach $0$. Therefore, for large enough $m$, we have
    \begin{align*}
        \entropy(G) \approx \log Z = \log \Big ( \sum_{\x\in\val(\X)} \sigma(w_0 + \sum_{i=1}^{n} w_i \cdot x_i) \Big ) = \log \Big ( \sum_{\x\in\val(\X)} \sigma(w_0 + \sum_{i=1}^{n} w_i \cdot x_i) \Big ).
    \end{align*}
For each $S \subseteq [n]$, we define $\mathrm{weight}(S) := -\frac{m}{2} - mc + m (\sum_{i \in S} k_i)$. Then, 
    \begin{align*}
        \exp(\entropy(G)) & \approx \sum_{\x\in\val(\X)} \sigma( -\frac{m}{2} - mc + m (\sum_{i \in [n]} k_i x_i)) \\
        & = \sum_{\x\in\val(\X)} \sigma( -\frac{m}{2} - mc + m (\sum_{i:x_i=1} k_i)) \\
        & = \sum_{S \subseteq [n]} \sigma( \mathrm{weight} (S) ) \\
        & = \frac{1}{2} \sum_{S \subseteq [n]} \big ( \sigma( \mathrm{weight} (S) ) + \sigma( \mathrm{weight} (\bar{S}) ) \big ).
    \end{align*}
If $S$ is a solution to \textsc{\#NUMPAR}, then
    \begin{align*}
        \sigma( \mathrm{weight} (S) ) + \sigma( \mathrm{weight} (\bar{S}) ) = 2 \sigma(-m/2).
    \end{align*}
Othervise, one of $\mathrm{weight} (S)$ and $\mathrm{weight} (\bar{S})$ is $\geq m/2$ and the other is $\leq -3m/2$, and hence
    \begin{align*}
        \sigma(m/2) \leq \sigma( \mathrm{weight} (S) ) + \sigma( \mathrm{weight} (\bar{S}) ) \leq 1 + \sigma(-3m/2).
    \end{align*}
For a large enough $m$ such that $2 \sigma(-m/2) < \epsilon$ and $1-\sigma(m/2) < \epsilon$, we have
    \begin{align*}
        S \in P: & \qquad 0 \leq \sigma( \mathrm{weight} (S) ) + \sigma( \mathrm{weight} (\bar{S}) ) \leq \epsilon, \\
        S \not\in P: & \quad 1-\epsilon \leq \sigma( \mathrm{weight} (S) ) + \sigma( \mathrm{weight} (\bar{S}) ) \leq 1 + \epsilon.
    \end{align*}
Therefore, we have
    \begin{gather*}
        \frac{2^n - \abs{P}}{2} (1-\epsilon) \leq \exp(\entropy(G)) \leq \frac{\abs{P}}{2} \epsilon + \frac{2^n - \abs{P}}{2} (1+\epsilon) \\
        \abs{P} \geq 2^n - \frac{2 \exp(\entropy(G))}{1-\epsilon} \\
        \abs{P} \leq 2^n (1+\epsilon) - 2 \exp(\entropy(G))
    \end{gather*}
This gives a lower and upper bound for $\abs{P}$. For small enough $\epsilon$ (governed by large enough $m$), the difference between the lower and upper bound is less than $1$, and hence $\abs{P}$ can be uniquely determined, which proves the theorem. 

\subsection{Proof of Theorem~\ref{thm:ent-reg-alg-correctness}}
\label{sec:proof-ent-reg-alg}

First note that according to \cref{lem:ent-non-concave}, \cref{eq:det-pc-ent-reg-mle} is not a convex optimization problem. The key idea of \cref{alg:ent-reg} is to propose a set of \emph{surrogate objective} functions, and maximize the objective function \cref{eq:det-pc-ent-reg-mle} by iteratively maximizing the surrogate objective. Concretely, we show the monotonic convergence property of \cref{alg:ent-reg} by checking the correctness of the following three statements: \newline
$\bullet$ \textbf{Statement \#1:} The surrogate objective is easy to maximize as it is a concave function w.r.t. the parameters. \newline
$\bullet$ \textbf{Statement \#2:} The surrogate objective is consistent with the original objective \cref{eq:det-pc-ent-reg-mle}. That is, whenever a set of surrogate objectives are improved, the true objective is also improved. \newline
$\bullet$ \textbf{Statement \#3:} The surrogate objectives can always be improved unless the original objective \cref{eq:det-pc-ent-reg-mle} has zero first-order derivative. \newline
$\bullet$ \textbf{Statement \#4:} Solving \cref{eq:ent-reg-eqs} is equivalent to maximizing the surrogate objective.

Before verifying the statements, we first formally define the surrogate. Denote $\entropy(\p_n;\params)$ as the entropy of the PC rooted at $n$ and with parameters $\params$; the top-down probability of $n$, denoted $\Pr_{n_r}(n)$, is recursively defined as follows:
    \begin{align}
        \Pr_{n_r}(n) :=
        \begin{cases}
            1 & \text{if}~$n$~\text{is the root node}~n_r, \\
            \sum_{m\in\pa(n)} \Pr_{n_r}(m) & \text{if}~$n$~\text{is a sum unit}, \\
            \sum_{m\in\pa(n)} \theta_{m,n} \cdot \Pr_{n_r}(m) & \text{if}~$n$~\text{is a product unit}.
        \end{cases}
        \label{eq:def-pr-n}
    \end{align}

Given a set of reference parameters $\params^{\text{ref}}$, we define the surrogate objective w.r.t. parameter $\theta_{n,c}$ as
    \begin{align*}
        \calL_{\text{surr}} (\theta_{n,c}; \params^{\text{ref}}) \!:=\! & \underbrace{\frac{1}{N} \sum_{i = 1}^{N} \log \p (\x^{(i)}; \params^{\text{ref}}\backslash\{\theta^{\text{ref}}_{n,c}\}, \theta_{n,c})}_{\text{Term 1}} \\
        & + \underbrace{\tau \!\cdot\! \Pr_{n_r}(n;\params^{\text{ref}}) \!\cdot\! \Big ( \!\! -\!\theta_{n,c} \log \theta_{n,c} \!+\! \theta_{n,c} \!\cdot\! \entropy(\p_c; \params^{\text{ref}}) \Big )}_{\text{Term 2}}.
        \numberthis\label{eq:def-surrogate-loss}
    \end{align*}
    
Given parameters $\params^{\text{old}}$, we now describe an update procedure to obtain a set of new parameters $\params^{\text{new}}$. 

\boldparagraph{Parameter update procedure}
We start with an empty set of parameters $\params^{\text{update}} := \params^{\text{old}}$ and iteratively update its entries with updated parameters $\theta^{\text{new}}_{n,c}$. For every sum unit $n$ traversed in pre-order, we update the parameters $\{\theta_{n,c} \mid c\in\ch(n)\}$ by maximizing the sum of surrogate objectives:
    \begin{align}
        \sum_{c\in\ch(n)} \calL_{\text{surr}} (\theta_{n,c}; \params^{\text{update}}).
        \label{eq:surrogate-ent-loss}
    \end{align}
After solving the above equation, the updated parameters $\{\theta_{n,c} \mid c\in\ch(n)\}$ replace the corresponding original parameters in $\params^{\text{update}}$. As we will proceed to show in statement \#4, maximizing \cref{eq:surrogate-ent-loss} is done in Lines 7 to 7 in \cref{alg:ent-reg}.

Given the formal definition of the surrogate objective and the corresponding update process, we re-state the three statements and prove their validity in the following.

$\bullet$ \textbf{Statement \#1:} The surrogate objective \cref{eq:surrogate-ent-loss} is concave w.r.t. parameters $\{\theta_{n,c} \mid c\in\ch(n)\}$.

\begin{proof}
This statement can be proved by showing that $\forall (n,c), \forall \params$, $\calL_{\text{surr}} (\theta_{n,c}; \params)$ is concave. Specifically, according to \cref{lem:mle-concavity}, the first term of \cref{eq:def-surrogate-loss} is concave; the second term of \cref{eq:def-surrogate-loss} is concave since (i) $-x\log x$ is concave w.r.t. $x$, and (ii) $\Pr_{n_r}(n;\params^{\text{ref}})$ and $\entropy(\p_c;\params^{\text{ref}})$ are independent of $\{\theta_{n,c'} \mid c'\in\ch(n)\}$.
\end{proof}

$\bullet$ \textbf{Statement \#2:} For any sum unit $n$ and any parameters $\params$, if we update $n$'s parameters (\ie $\{\theta_{n,c} \mid c\in\ch(n)\}$) by maximizing \cref{eq:surrogate-ent-loss}, the true objective \cref{eq:det-pc-ent-reg-mle} will also improve.

\begin{proof}
Consider updating the parameters correspond to sum unit $n$ (\ie $\{\theta_{n,c} \mid c\in\ch(n)\}$) by maximizing \cref{eq:surrogate-ent-loss}. We can re-arrange the entropy $\entropy(\p_{n_r})$ as follows:
    \begin{align*}
        \entropy(\p_{n_r}) & \overset{(a)}{=} - \sum_{(n',c')\in\mathtt{edges}(\p_{n_r})} \Pr_{n_r}(n') \cdot \theta_{n',c'} \log \theta_{n',c'} \\
        & = - \sum_{(n',c')\in\mathtt{edges}(\p_{n})} \Pr_{n_r}(n') \cdot \theta_{n',c'} \log \theta_{n',c'} + \mathrm{const} \\
        & = - \sum_{(n',c')\in\mathtt{edges}(\p_{n})} \Big ( \sum_{m\in\pa(n)} \Pr_{n_r}(m) \Big ) \cdot \Pr_{n}(n') \cdot \theta_{n',c'} \log \theta_{n',c'} + \mathrm{const} \\
        & = - \sum_{(n',c')\in\mathtt{edges}(\p_{n})} \Pr_{n_r}(n) \cdot \Pr_{n}(n') \cdot \theta_{n',c'} \log \theta_{n',c'} + \mathrm{const} \\
        & = \Pr_{n_r}(n) \cdot \entropy(\p_{n}) + \mathrm{const} \\
        & \overset{(b)}{=} \Pr_{n_r}(n) \cdot \sum_{c\in\ch(n)} \Big ( \!\! -\!\theta_{n,c} \log \theta_{n,c} \!+\! \theta_{n,c} \!\cdot\! \entropy(\p_c) \Big ) + \mathrm{const},
    \end{align*}
\noindent where $\mathrm{const}$ denotes terms that do not depend on $\{\theta_{n,c'} \mid c'\in\ch(n)\}$; $(a)$ and $(b)$ directly apply \cref{lem:entropy-decomposition} and \cref{lem:det-pc-ent-decom}, respectively.

Thus, the true objective \cref{eq:det-pc-ent-reg-mle} can be written as follows:
    \begin{align*}
        & \frac{1}{N} \sum_{i=1}^{N} \log \p(\x^{(i)}) + \tau \cdot \entropy(\p) \\
        = & \frac{1}{N} \sum_{i=1}^{N} \log \p(\x^{(i)}) + \tau \cdot \Pr_{n_r}(n) \cdot \sum_{c\in\ch(n)} \Big ( \!\! -\!\theta_{n,c} \log \theta_{n,c} \!+\! \theta_{n,c} \!\cdot\! \entropy(\p_c) \Big ) + \mathrm{const}
        \numberthis\label{eq:thm2-proof-eq1}
    \end{align*}
Compare \cref{eq:thm2-proof-eq1} and \cref{eq:def-surrogate-loss}, we can see that they only differs in some constant terms. Therefore, maximizing \cref{eq:def-surrogate-loss} w.r.t. $\{\theta_{n,c'} \mid c'\in\ch(n)\}$ will lead to an increase in the true objective \cref{eq:det-pc-ent-reg-mle}.
\end{proof}

$\bullet$ \textbf{Statement \#3:} The surrogate objectives can always be improved unless the original objective \cref{eq:det-pc-ent-reg-mle} has zero first-order derivative. 

\begin{proof}
Recall from \cref{eq:thm2-proof-eq1} that for any sum unit $n$, the true objective \cref{eq:det-pc-ent-reg-mle} can be written as the sum of \cref{eq:def-surrogate-loss} and terms that are independent with the parameters of $n$ (\ie $\{\theta_{n,c'} \mid c'\in\ch(n)\}$). Therefore, the true objective can always be improved by maximizing the surrogate objective \cref{eq:surrogate-ent-loss} as long as the true objective has non-zero first-order derivative w.r.t. the parameters.
\end{proof}

$\bullet$ \textbf{Statement \#4:} Solving \cref{eq:ent-reg-eqs} is equivalent to maximizing the surrogate objective.

\begin{proof}
We want to maximize the surrogate objective given the assumption that the parameters w.r.t. a sum unit sum up to 1:
    \begin{align}
        \maximize_{\theta_{n,c}} \calL_{\text{surr}} (\theta_{n,c}; \params^{\text{ref}}), \; \text{such~that}~\sum_{c\in\ch(n)} \theta_{n,c} = 1.
        \label{eq:proof-thm2-1}
    \end{align}
Since the surrogate objective $\calL_{\text{surr}} (\theta_{n,c}; \params^{\text{ref}})$ is concave, maximizing the surrogate objective is equivalent to finding its stationary point. Specifically, we solve \cref{eq:proof-thm2-1} with the Lagrange multiplier method (variable $\lambda$ corresponds to the constraint):
    \begin{align*}
        \maximize_{\theta_{n,c}} \minimize_{\lambda} \calL_{\text{surr}} (\theta_{n,c}; \params^{\text{ref}}) - \lambda (1 - \sum_{c\in\ch(n)} \theta_{n,c})    
    \end{align*}
Its KKT conditions can be written as:
    \begin{align*}
        \begin{cases}
            \frac{\flow_{n,c_i}(\data)}{\abs{\data} \cdot \theta_{n,c_i}} - \tau \cdot \Pr_{n_r}(n;\params^{\text{ref}}) (\log \theta_{n,c_i} + 1 + \entropy(\p_{c_i}; \params^{\text{ref}})) + \lambda = 0 & (\forall 1 \leq i \leq \abs{\ch(n)}), \\
            \sum_{c\in\ch(n)} \theta_{n,c} = 1.
        \end{cases}
    \end{align*}
It is easy to verify that the above equation is equivalent to \cref{eq:ent-reg-eqs} by substituting the definitions in Lines 7-8 in \cref{alg:ent-reg}.
    
\end{proof}

Therefore, by following the parameter update procedure, we can always make progress since the surrogate objective is concave (statement \#1) and the true objective improves as long as the surrogate objective increases (statement \#2). Finally, the learning procedure will not terminate unless a local maximum is achieved (statement \#3).

\subsection{Correctness of Algorithms~\ref{alg:forward-pass}~and~\ref{alg:backward-pass}}
\label{sec:correctness-f-b}

The correctness of \cref{alg:forward-pass,alg:backward-pass} can be justified directly by the proof of \cref{thm:data-softening-alg}. Specifically, since with $\beta = 1$, the softened dataset $\data_{\beta}$ is equivalent to $\data$, we can use the proof in \cref{sec:proof-data-soft-alg} and set $\beta = 1$ (the proof holds for any $\beta \in (0.5,1]$).

\subsection{Proof of Proposition~\ref{prop:hardness-ent-reg}}
\label{sec:proof-hardness-ent-reg}

The first statement (\ie \cref{eq:det-pc-ent-reg-mle}) could be non-concave) is proved in \cref{lem:ent-non-concave}. The second statement (\ie \cref{eq:det-pc-ent-reg-mle} could have multiple local maximas) is proved in \cref{lem:ent-multimodal}.

\begin{figure}[t]
\begin{minipage}[t]{0.50\textwidth}
\begin{algorithm}[H]
\caption{Forward pass (expected flows)}
\label{alg:nondet-forward-pass}
{\fontsize{9}{9} \selectfont
\begin{algorithmic}[1]

\STATE {\bfseries Input:} A non-deterministic PC $\p$; sample $\x$

\STATE {\bfseries Output:} $\mathtt{value}[n] \!\!:=\!\! (\x\!\in\!\supp(n))$ for each unit $n$

\ForEach
\FOR{\tikzmarknode{a1}{} $n$ traversed in postorder}
\STATE \textbf{if} $n$ \textbf{isa} input unit \textbf{then} $\mathtt{value}[n]\!\leftarrow\!f_n(\x)$
\STATE \textbf{elif} \tikzmarknode{a2}{} $n$ \textbf{isa} product unit \textbf{then}
\STATE \hspace{1em} $\mathtt{value}[n]\!\leftarrow\!\prod_{c\in\ch(n)} \mathtt{value}[c]$
\STATE \textbf{else} \tikzmarknode{a3}{} \slash\slash $n$ is a sum unit
\STATE \hspace{1em} $\mathtt{value}[n]\!\leftarrow\!\sum_{c\in\ch(n)} \theta_{n,c} \cdot \mathtt{value}[c]$
\ENDFOR
\ForOnly
\end{algorithmic}
}
\end{algorithm}
\end{minipage}
\hfill
\begin{minipage}[t]{0.49\textwidth}
\begin{algorithm}[H]
\caption{Backward pass (expected flows)}
\label{alg:nondet-backward-pass}
{\fontsize{9}{9} \selectfont
\begin{algorithmic}[1]

\STATE {\bfseries Input:} A non-deterministic PC $\p$; $\forall n, \mathtt{value}[n]$

\STATE {\bfseries Output:} $\mathtt{eflow}[n,c] \!:=\! \expectation_{\z\in\p_{c}(\cdot\mid\x;\params)} ((\x,\z)\!\in\!(\context_n \!\cap\! \context_c))$ for each pair $(n,c)$, where $n$ is a sum unit and $c\!\in\!\ch(n)$

\STATE $\forall n, \mathtt{context}[n]\!\leftarrow\!0$; $\mathtt{context}[n_r]\!\leftarrow\! \mathtt{value}[n_r]$

\ForEach
\FOR{\tikzmarknode{b1}{} sum unit $n$ traversed in preorder}
\NoDo
\FOR{\tikzmarknode{b2}{} $m \in \pa(n)$ \textbf{do} $\;$ (denote $g\!\leftarrow\!\pa(m)$)}
\STATE $\mathtt{f} \leftarrow \frac{\mathtt{value}[m]}{\mathtt{value}[g]} \cdot \mathtt{context}[g] \cdot \theta_{g,m}$
\STATE $\mathtt{context}[n] \pluseq \mathtt{f}; \quad \mathtt{flow}[g,m] = \mathtt{f}$
\ENDFOR
\ReDo
\ENDFOR
\ForOnly

\end{algorithmic}
}
\end{algorithm}
\end{minipage}
\begin{tikzpicture}[overlay,remember picture]
   \draw[black,line width=0.6pt] ([xshift=-32pt,yshift=-3pt]a1.west) -- ([xshift=-32pt,yshift=-52pt]a1.west) -- ([xshift=-28pt,yshift=-52pt]a1.west);
   \draw[black,line width=0.6pt] ([xshift=-13pt,yshift=-3pt]a2.west) -- ([xshift=-13pt,yshift=-12pt]a2.west) -- ([xshift=-9pt,yshift=-12pt]a2.west);
   \draw[black,line width=0.6pt] ([xshift=-14.4pt,yshift=-3pt]a3.west) -- ([xshift=-14.4pt,yshift=-12pt]a3.west) -- ([xshift=-10.4pt,yshift=-12pt]a3.west);
   \draw[black,line width=0.6pt] ([xshift=-32pt,yshift=-3pt]b1.west) -- ([xshift=-32pt,yshift=-37pt]b1.west) -- ([xshift=-28pt,yshift=-37pt]b1.west);
   \draw[black,line width=0.6pt] ([xshift=-32pt,yshift=-3pt]b2.west) -- ([xshift=-32pt,yshift=-27pt]b2.west) -- ([xshift=-28pt,yshift=-27pt]b2.west);
\end{tikzpicture}
\end{figure}

\section{Method or Experiment Details}

\subsection{Soften non-boolean datasets}
\label{sec:softreg-non-boolean}

As a direct extension of softening boolean datasets, datasets with categorical variables can be similarly softened. Suppose $X$ is a categorical variable with $k$ categories. For an assignment $x = j$, we can soften it as follows
    \begin{align*}
        \begin{cases}
            \Pr(x = i) = \frac{1-\beta}{k} & (i \neq j), \\
            \Pr(x = j) = \beta.
        \end{cases}
    \end{align*}
To compute the flow $F_{n,c} (\data_{\beta})$ w.r.t. a softened categorical dataset, we can again adopt \cref{alg:forward-pass,alg:backward-pass} by choosing
    \begin{align*}
        f_{n} (\x) = \beta \!\cdot\! \indicator{\x\in\supp(n)} + \frac{1-\beta}{k} \!\cdot\! \indicator{\x\not\in\supp(n)}.
    \end{align*}

\subsection{Solving Equation~\ref{eq:ent-reg-eqs}}
\label{sec:solving-entreg-eqs}

Denote $\gamma_{c_i} \!:=\! \mathtt{entropy}[c_i]$, our goal is to solve the following set of equations:
    \begin{align*}
        \begin{cases}
            d_i e^{-\varphi_{n,c_i}} - b \cdot \varphi_{n,c_i} + b \cdot \gamma_{c_i} = y \quad (\forall i \in \{1,\dots,\abs{\ch(n)}\}), \\
            \sum_{i=1}^{\abs{\ch(n)}} e^{\varphi_{n,c_i}} = 1.
        \end{cases}
    \end{align*}
We break down the problem by iteratively solve for $\{\varphi_{n,c_i}\}_{i=1}^{\abs{\ch(n)}}$ and $y$, respectively.

$\bullet$ Solve for $y$. Given variables $\{\varphi_{n,c_i}\}_{i=1}^{\abs{\ch(n)}}$, we update $y$ as
    \begin{align*}
        y = \frac{1}{\abs{\ch(n)}} \sum_{i=1}^{\abs{\ch(n)}} d_i e^{-\varphi_{n,c_i}} - b \cdot \varphi_{n,c_i} + b \cdot \gamma_{c_i}.
    \end{align*}
    
$\bullet$ Solve for $\{\varphi_{n,c_i}\}_{i=1}^{\abs{\ch(n)}}$. Given $y$, we first update each $\varphi_{n,c_i}$ individually by solving the equation 
    \begin{align*}
        d_i e^{-\varphi_{n,c_i}} - b \cdot \varphi_{n,c_i} + b \cdot \gamma_{c_i} = y.
    \end{align*}
Specifically, this is done by iterative Newton method update:
    \begin{align*}
        \varphi_{n,c_i} \pluseq \frac{\frac{d_i}{\varphi_{n,c_i}} + b \cdot (\gamma_{c_i} - \varphi_{n,c_i}) + y}{\frac{d_i}{\varphi_{n,c_i}} + b}
    \end{align*}
After one Newton method update step for every parameter in $\{\varphi_{n,c_i}\}_{i=1}^{\abs{\ch(n)}}$, we enforce the constraint $\sum_{i=1}^{\abs{\ch(n)}} e^{\varphi_{n,c_i}} = 1$ by
    \begin{align*}
        \varphi_{n,c_i} \minuseq \log \Big ( \sum_{i=1}^{\abs{\ch(n)}} e^{\varphi_{n,c_i}} \Big ).
    \end{align*}

\subsection{Details of the Experiments on Deterministic PCs}
\label{sec:detail-det-pc-exps}

\boldparagraph{PC structures}
For each dataset, we adopt 16 PCs by running Strudel \citep{dang2020strudel} for $\{1000, 1200, 1400, \dots, 4000\}$ iterations except for the dataset ``dna'', which we ran Strudel for $\{50, 100, 150, \dots, 800\}$ iterations since the learning algorithm takes significantly longer for this dataset.

\boldparagraph{Hyperparameters}
We always perform hyperparameter search using the validation set, and report the final performance on the test set. Whenever we use data softening or entropy regularization, we also add pseudocount $\alpha \!=\! 1$ since it yields better performance.

\textbf{Server specifications} All our experiments were run on a server with 72 CPUs, 512G Memory, and 2 TITAN RTX GPUs.

\subsection{Details of the Experiments on Non-Deterministic PCs}
\label{sec:nondet-exps-details}

\boldparagraph{The HCLT structure}
For the experiments on the twenty datasets, we set the hidden size of the HCLT structure as $12$, \ie every latent variable $Z$ is a categorical variable with $12$ categories. Additionally, following \cite{dang2020strudel,liang2017learning}, we learn a mixture of $4$ HCLTs to achieve better performance. 
For the protein sequence dataset, we adopted a mixture of $2$ HCLTs with hidden size $32$.

\boldparagraph{Detailed results} 
As an extension of \cref{tab:20datasets}, \cref{tab:20datasets-extend} provides the average test set log-likelihood for all adopted baselines.

\begin{table}[t]
    \caption{Full results on the 20 density estimation benchmarks. As an extension of \cref{tab:20datasets}, we report the average test-set log-likelihood of all baselines: Strudel \citep{dang2020strudel}, LearnPSDD \citep{liang2017learning}, EinSumNet \citep{peharz2020einsum}, LearnSPN \citep{gens2013learning}, ID-SPN \citep{rooshenas2014learning}, and RAT-SPN \citep{peharz2020random}.}
    \label{tab:20datasets-extend}
    \centering
    \scalebox{1.0}{
    {\renewcommand{\arraystretch}{0.8}
    \begin{tabular}{lccccccc}
        \toprule
        Dataset & HCLT & EiNet & LearnSPN & ID-SPN & RAT-SPN & Strudel & LearnPSDD \\
        \midrule
        accidents & -26.78 & -35.59 & -40.50 & -26.98 & -35.48 & -29.46 & -28.29 \\
        ad & -16.04 & -26.27 & -19.73 & -19.00 & -48.47 & -16.52 & -20.13 \\
        baudio & -39.77 & -39.87 & -40.53 & -39.79 & -39.95 & -42.26 & -41.51 \\
        bbc & -250.07 & -248.33 & -250.68 & -248.93 & -252.13 & -258.96 & -260.24 \\
        bnetflix & -56.28 & -56.54 & -57.32 & -56.36 & -56.85 & -58.68 & -58.53 \\
        book & -33.84 & -34.73 & -35.88 & -34.14 & -34.68 & -35.77 & -36.06 \\
        c20ng & -151.92 & -153.93 & -155.92 & -151.47 & -152.06 & -160.77 & -160.43 \\
        cr52 & -84.67 & -87.36 & -85.06 & -83.35 & -87.36 & -92.38 & -93.30 \\
        cwebkb & -153.18 & -157.28 & -158.20 & -151.84 & -157.53 & -160.50 & -161.42 \\
        dna & -79.33 & -96.08 & -82.52 & -81.21 & -97.23 & -87.10 & -83.02 \\
        jester & -52.45 & -52.56 & -75.98 & -52.86 & -52.97 & -55.30 & -54.63 \\
        kdd & -2.18 & -2.18 & -2.18 & -2.13 & -2.12 & -2.17 & -2.17 \\
        kosarek & -10.66 & -11.02 & -10.98 & -10.60 & -10.88 & -10.98 & -10.99 \\
        msnbc & -6.05 & -6.11 & -6.11 & -6.04 & -6.03 & -6.05 & -6.04 \\
        msweb & -9.90 & -10.02 & -10.25 & -9.73 & -10.11 & -10.19 & -9.93 \\
        nltcs & -6.00 & -6.01 & -6.11 & -6.02 & -6.01 & -6.06 & -6.03 \\
        plants & -14.31 & -13.67 & -12.97 & -12.54 & -13.43 & -13.72 & -13.49 \\
        pumbs* & -23.32 & -31.95 & -24.78 & -22.40 & -32.53 & -25.28 & -25.40 \\
        tmovie & -50.69 & -51.70 & -52.48 & -51.51 & -53.63 & -59.47 & -55.41 \\
        tretail & -10.84 & -10.91 & -11.04 & -10.85 & -10.91 & -10.90 & -10.92 \\
        \bottomrule
    \end{tabular}}}
\end{table}

\end{document}